\ifdefined\pdfminorversion\pdfminorversion=7\relax\fi
\documentclass[10pt,twocolumn,letterpaper]{article}
\usepackage[pagenumbers]{cvpr}
\usepackage[T1]{fontenc}
\usepackage{array,tabularx,longtable,enumitem}
\usepackage{multirow}
\usepackage{float}
\floatstyle{ruled}
\newfloat{algorithm}{tbp}{loa}
\floatname{algorithm}{Algorithm}
\usepackage{microtype}
\usepackage{tikz}
\usepackage{tcolorbox}
\usetikzlibrary{arrows.meta,positioning}
\tcbuselibrary{listings,skins,breakable}

\definecolor{promptbackground}{HTML}{F7F9FC}
\definecolor{promptborder}{HTML}{A9BACB}
\definecolor{prompttitlebackground}{HTML}{E9EFF5}
\definecolor{prompttitletext}{HTML}{29445E}
\definecolor{perceptioncolor}{HTML}{2A4E7E}
\definecolor{reasoningcolor}{HTML}{2E6B45}
\definecolor{reportingcolor}{HTML}{8A5A20}
\definecolor{gaingreen}{HTML}{1B7F3B}

\newtcblisting{promptbox}[1]{
    enhanced, breakable, listing only,
    title={#1},
    colback=promptbackground, colframe=promptborder,
    colbacktitle=prompttitlebackground, coltitle=prompttitletext,
    fonttitle=\sffamily\bfseries\fontsize{9}{11}\selectfont,
    boxrule=0.45pt, arc=2pt, outer arc=2pt, boxsep=0pt,
    left=8pt, right=8pt, top=5pt, bottom=5pt,
    toptitle=4pt, bottomtitle=4pt,
    before skip=6pt, after skip=6pt,
    listing options={
        basicstyle=\ttfamily\fontsize{8}{10}\selectfont,
        columns=fullflexible, keepspaces=true,
        breaklines=true, breakatwhitespace=false, breakindent=8pt,
        showstringspaces=false, showspaces=false, showtabs=false,
        tabsize=2, aboveskip=0pt, belowskip=0pt,
        escapeinside={(*@}{@*)}
    }
}

\definecolor{cvprblue}{rgb}{0.21,0.49,0.74}
\usepackage[pagebackref,breaklinks,colorlinks,allcolors=cvprblue]{hyperref}

\title{Decoupling Perception from Generation for Agentic 3D CT Report Generation}

\author{%
    Chenyu Wang \quad Mingyu Chen \quad Wenchao Li \quad Kayhan Batmanghelich \\
    Boston University, Boston, MA, USA \\
    \texttt{\{chyuwang, mingyuc, wenchao, batman\}@bu.edu}
}

\hypersetup{
    pdftitle={Decoupling Perception from Generation for Agentic 3D CT Report Generation},
    pdfauthor={Chenyu Wang, Mingyu Chen, Wenchao Li, Kayhan Batmanghelich}
}

\begin{document}
\maketitle
\makeatletter
\let\cvprsavedtextbottom\@textbottom
\def\@textbottom{%
  \ifnum\c@page>3\relax
    \ifnum\c@page<6\relax
      \vskip\z@\@plus.0001fil
    \else\cvprsavedtextbottom\fi
  \else\cvprsavedtextbottom\fi}
\makeatother
\begin{abstract}
Chest CT reports repeatedly assess the same anatomical regions for findings and their attributes, such as whether a lung nodule is present and where it is located. These assessments suggest learning to answer visual questions directly, rather than coupling perception to open-ended text generation. Yet current CT report generation models largely follow generic vision--language modeling, which may miss subtle abnormalities when training data are limited. We therefore formulate report generation as a set of visual question-answering decisions coordinated by a large language model (LLM) agent. This decouples visual perception from reasoning and report writing: perception can be learned as a discriminative, non-generative task, while the LLM focuses on reasoning over patient context and posing relevant visual queries. We introduce Rad-VL-JEPA, which unifies report retrieval, finding recognition, and fine-grained attribute characterization through vision--language joint-embedding prediction. Building on this perception module, an agent trained with reinforcement learning (RL) acquires patient-specific visual evidence, while a separate editor incorporates the evidence into the final report, further separating evidence acquisition from report realization. Experiments show that Rad-VL-JEPA achieves strong performance on fine-grained CT understanding tasks with lower amortized per-query inference time than a generative VLM baseline trained with the same CT encoder and QA data. The resulting reporting system outperforms strong VLM-based and agentic baselines.
\end{abstract}


\section{Introduction}

Automated chest CT reporting aims to describe abnormalities and their attributes accurately from a volumetric scan. Reliable automation could reduce the burden of reviewing hundreds of slices and documenting the findings~\citep{roschewitz2026radagent}. The challenge extends beyond recognizing whether an abnormality is present: a useful report must also specify its location, size, and severity~\citep{yuan2026ct}. This requires both fine-grained visual perception and faithful expression of the resulting evidence in text.

Most current CT report generators couple a visual encoder with a language model trained to generate reports token by token~\citep{hamamci2026generalist,lin2026omnict,myronenko2026nvreasonct}. This objective requires models to learn visual perception while also modeling lexical variation, and they can miss subtle abnormalities when training data are limited~\citep{hamamci2026generalist,chen2026vl}. Optimizing this objective can also encourage models to rely on language priors that bypass visual evidence, producing fluent reports that omit or mischaracterize fine-grained attributes~\citep{cheng2026rethinking,wang2026enhancing}. To identify and correct such errors, recent work adopts agentic workflows in which an LLM calls visual tools to collect preliminary visual findings before writing the report~\citep{lin2026march,roschewitz2026radagent}. However, these tools often specialize in anatomical regions or finding presence~\citep{mao2026ct,roschewitz2026radagent}, with finer questions delegated to generalist VLMs~\citep{hamamci2026generalist}. This leaves two practical gaps. First, overall report scores do not establish how reliably a visual tool answers each finding--attribute question, making its strengths and failures difficult to isolate~\citep{yuan2026ct}. Second, repeatedly invoking a large generative backbone incurs token-by-token decoding for each answer, increasing the latency of evidence acquisition~\citep{chen2026vl,roschewitz2026radagent}.

We observe that chest CT's bounded anatomical field of view gives reporting a recurring structure: reports assess a shared set of regions, findings, and attributes~\citep{zhang2025development,yuan2026ct}. This structure supports a clinical question set~\citep{agrawal2025pillar} and query-specific answer banks, so that each visual answer can be inspected and evaluated independently of the final report. Moreover, answers with the same clinical meaning may differ in wording. Learning their semantic representations can reduce the need to model this lexical variation~\citep{chen2026vl}. These observations motivate decoupling visual perception from reasoning and writing: perception learns to answer fine-grained questions directly, reasoning determines which questions matter for the current patient, and writing incorporates the resulting evidence through minimal revisions to the draft.

\begin{figure*}[t]
    \centering
    \includegraphics[width=\linewidth]{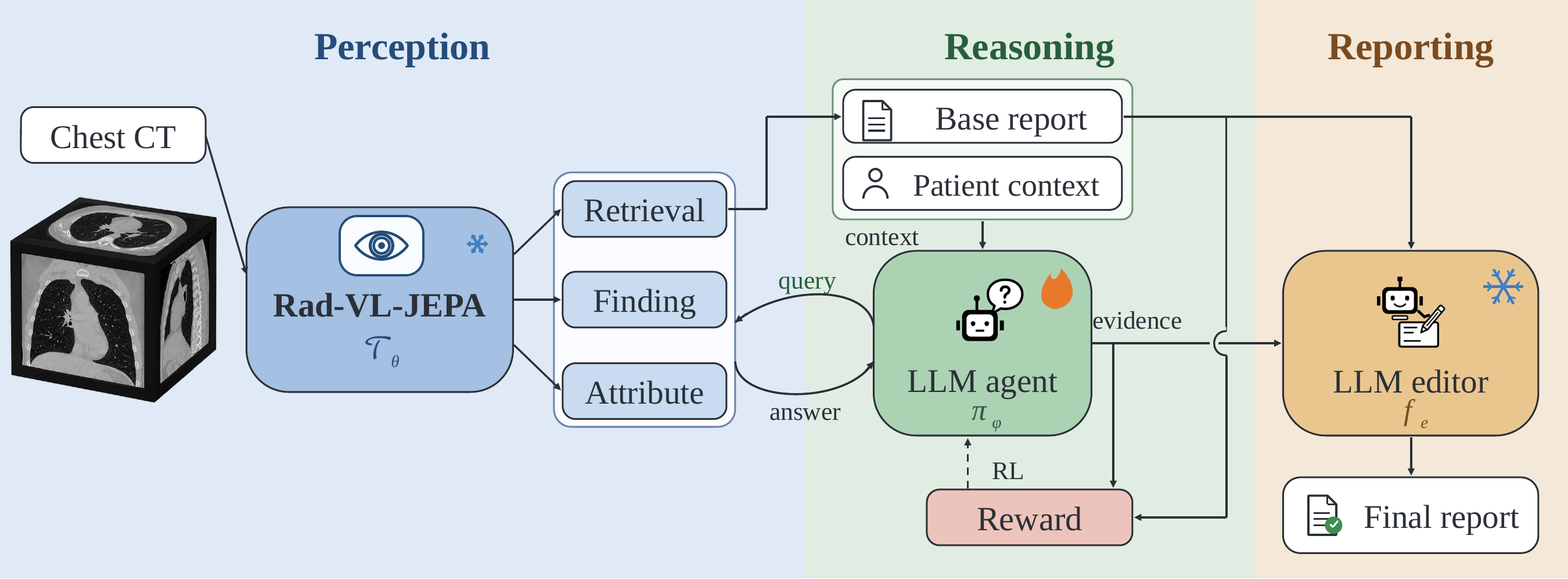}
\caption{\textbf{Decoupled agentic workflow for chest CT report generation.} The perception tool $\mathcal T_\theta$ (Rad-VL-JEPA) retrieves a draft from training-set reports of other patients and answers visual queries. The query policy $\pi_\phi$, trained by RL using the correction reward, iteratively acquires finding and attribute evidence conditioned on the draft, patient context, and previous tool responses. The LLM editor $f_e$ incorporates the evidence through minimal edits to produce the final report. During policy RL, $\mathcal T_\theta$ and $f_e$ remain frozen; only $\pi_\phi$ is updated.}
    \label{fig:overview}
\end{figure*}

To address these gaps, we introduce \textbf{Rad-VL-JEPA}, which decouples fine-grained visual perception from reasoning and report writing (Figure~\ref{fig:overview}). Built on a joint embedding predictive architecture (JEPA), it learns query-conditioned answer representations in latent space, emphasizing clinical semantics while reducing the need to model lexical variability~\citep{chen2026vl}. We further unify report retrieval, finding recognition, attribute characterization, and longitudinal diagnosis through a clinical question set and query-specific answer banks, making each prediction directly evaluable. This non-generative prediction design avoids token-by-token decoding during answer prediction for each query. Under a fixed query budget, we train an agent with reinforcement learning (RL) to select visual queries using patient history and cross-patient report context. A separate LLM editor integrates the acquired evidence through minimal edits, preserving unedited text and avoiding unsupported or unnecessary changes~\citep{gao2023rarr,huang2024fineradscore}.

Our contributions are as follows:
\begin{itemize}[leftmargin=*, topsep=2pt, itemsep=2pt, parsep=0pt]
    \item We reformulate chest CT report generation as coordinated visual question answering. Motivated by the recurring structure of anatomical regions, findings, and attributes within the chest CT field of view, this formulation decouples visual perception from reasoning and report writing.
    \item We introduce Rad-VL-JEPA, a unified non-generative perception model for chest CT. It unifies report retrieval, finding recognition, and fine-grained attribute characterization through a shared embedding interface, with lower training cost and faster inference than a generative VLM baseline trained with the same CT encoder and QA data.
    \item We develop a CT report generation system that integrates Rad-VL-JEPA for perception, an RL-trained agent for context-aware evidence acquisition, and an LLM editor for evidence-guided reporting. The system improves clinical report accuracy over strong VLM-based and agentic baselines.
\end{itemize}

\section{Related Work}

\subsection{Vision--Language Modeling for CT Report Generation}

Vision--language approaches to CT reporting follow two main strategies. Generative VLMs translate volumetric features into report text or answers to visual questions, as exemplified by CT-CHAT~\citep{hamamci2026generalist}. The generative objective couples visual perception with learning how observations are phrased; reliance on language priors can yield fluent descriptions that overlook abnormalities or assign incorrect attributes~\citep{cheng2026rethinking,wang2026enhancing}. IP-CRR adaptively queries existing radiology reports for interpretable diagnosis, but does not acquire evidence directly from images~\citep{ge2025ipcrr}. Agentic workflows distribute interpretation across tool calls before synthesizing a report~\citep{mao2026ct,lin2026march}. RadAgent, for example, combines finding classification with generative visual question answering to inspect an initial draft~\citep{roschewitz2026radagent}. Its perception capabilities depend on specialized predictors and VLM-based answers: detailed characterization still requires decoding text, and aggregate report scores cannot isolate errors in the acquired evidence~\citep{yuan2026ct}. Repeated decoding also adds latency to agent interaction~\citep{chen2026vl,roschewitz2026radagent}. We take the repeated clinical assessments associated with chest CT's anatomical coverage as the basis for a different decomposition. A shared discriminative model answers clinical visual questions, while an agent determines which evidence to request and an LLM editor incorporates it into a report. This makes visual decisions individually assessable and removes answer generation from the perception module.

\subsection{Visual Perception for CT Reporting}

CT visual perception aims to recognize abnormalities and characterize their attributes from volumetric scans. Existing approaches commonly learn visual representations through image--text alignment or visual self-supervision. CT-CLIP aligns CT volumes with reports for recognition and retrieval, while fVLM extends alignment to anatomical regions and their descriptions~\citep{hamamci2026generalist,shui2025fvlm}. Alternatively, Ker-VLJEPA uses a visual backbone pretrained through joint-embedding prediction on unlabeled CT scans~\citep{bumgardner2026curriculum}. However, these pretraining strategies may not adequately align fine-grained visual evidence with clinical descriptions. For detailed visual question answering, the learned representations are typically coupled with an autoregressive decoder, making visual answers dependent on text generation and introducing additional decoding cost. We adapt VL-JEPA~\citep{chen2026vl} with the pretrained Pillar-0~\citep{agrawal2025pillar} CT encoder and train a shared predictor that maps visual features and a textual clinical question to an answer embedding. The recurring regions, findings, and attributes assessed within chest CT's field of view motivate a clinical question bank and query-specific answer banks. These enable a unified perception model for report retrieval, finding recognition, and attribute characterization through candidate matching, without autoregressive answer generation.

\subsection{Reinforcement Learning for Report Generation}

Reinforcement learning (RL) has been applied to CT reporting to improve generated clinical outputs and train agents to use visual tools. NV-Reason-CT applies GRPO to a generative VLM, using rewards for abnormality prediction and report structure~\citep{myronenko2026nvreasonct}. RadAgent extends RL to an agentic workflow, jointly optimizing tool use and reporting through rewards for report quality, tool execution, and diagnostic workflow~\citep{roschewitz2026radagent}. However, report-quality feedback depends on both the evidence obtained and how it is expressed, which can obscure the contribution of evidence selection. We instead use RL to learn context-aware tool selection for visual question answering, directly rewarding queries that target missing or incorrect clinical information. The agent learns which findings and attributes to investigate based on clinical context and accumulated observations, making evidence acquisition the direct target of policy learning.

\section{Method}

Given a chest CT volume $x$ and patient context $c$, we generate a report $\hat r$ using the three components in Figure~\ref{fig:overview}. The perception agent retrieves a draft from other patients' training reports and answers visual questions. The reasoning agent selects queries using the draft, context, and previous question--answer pairs. The reporting agent edits the draft using acquired evidence.

\subsection{Perception: Rad-VL-JEPA}
\label{sec:perception_agent}

\paragraph{Architecture.}
Rad-VL-JEPA adapts VL-JEPA~\citep{chen2026vl} to map a CT volume $x$ and a clinical query $q$ to the embedding of the corresponding answer $a$.
The CT encoder $E_v$ extracts visual tokens, and a query-independent Perceiver resampler $R_\theta$ compresses them into a sequence of visual latents. The query is embedded using the language model's token embedding layer and concatenated with the projected visual latents. A predictor $P_\theta$, built from pretrained language-model layers adapted for bidirectional attention, jointly processes these inputs. Its hidden states are mean-pooled over valid visual and query positions and projected into the answer-embedding space. A target text encoder $E_a$ encodes the reference answer to provide the prediction target. The normalized prediction and target embeddings are given by
\begin{equation}
    \begin{aligned}
        f_\theta(x,q) &= \operatorname{norm}\!\left(P_\theta(R_\theta(E_v(x)),q)\right),\\
        e(a) &= \operatorname{norm}(E_a(a)),
    \end{aligned}
    \label{eq:answer_embeddings}
\end{equation}
where $\operatorname{norm}$ denotes $\ell_2$ normalization.

\paragraph{Training.}
Given training triplets $(x,q,a)\sim\mathcal D$, we align predicted and target answer embeddings by minimizing $\mathbb E_{(x,q,a)\sim\mathcal D}[1-f_\theta(x,q)^\top e(a)]$.
We optimize the resampler, predictor, and projection layers while keeping the CT encoder, query-token embeddings, and target text encoder frozen. We jointly train on shuffled VQA data spanning global reports, anatomical descriptions, finding presence, fine-grained attributes, and longitudinal reasoning. Appendix~\ref{app:jepa_implementation} gives architecture dimensions and optimization settings.

\paragraph{Inference.}
Rad-VL-JEPA scores candidates in a query-specific bank $\mathcal A(q)$ by cosine similarity. Report retrieval, attribute characterization, and longitudinal diagnosis return the highest-scoring candidate:
\begin{equation}
    \mathcal T_\theta(x,q)=\underset{a\in\mathcal A(q)}{\arg\max}\;f_\theta(x,q)^\top e(a).
    \label{eq:interface}
\end{equation}
The query and candidate bank determine the task. For report retrieval, a global report query ranks training-set reports, excluding those from the same patient. For binary finding recognition, $\mathcal T_\theta$ instead applies a finding-specific threshold to the \textsc{Yes}--\textsc{No} similarity margin, calibrated to maximize $F_1$ on the training partition. Calibration details are provided in Appendix~\ref{app:finding_calibration}. For attribute characterization, query-specific answer banks are built from training VQA answers (e.g., left lower lobe) and filtered by an LLM to remove off-topic candidates. Bank construction is detailed in Appendix~\ref{app:answer_banks}. For temporal characterization, queries additionally incorporate historical finding-presence labels, and the candidate states are \textsc{Refractory}, \textsc{Resolved}, \textsc{New}, and \textsc{No Abnormality}. For example, \textsc{New} denotes a finding present on the current scan but absent at the prior visit. These tasks share the same embedding-based scoring interface without autoregressive answer generation.

\subsection{Reasoning Agent: Adaptive Query Selection}
\label{sec:evidence_acquisition}

Effective query selection depends on both patient context and Rad-VL-JEPA's performance. We train a context-aware policy with reinforcement learning to select finding and attribute queries based on clinical context and previously acquired evidence.

\paragraph{Adaptive interaction.}
Given a CT volume $x$, the perception module retrieves the three highest-ranked training-set reports from other patients, sorted by the similarity in Eq.~\ref{eq:interface}. 
The highest-ranked report serves as the initial draft $r_0$. 
The agent selects queries from $\mathcal Q=\mathcal Q_F\cup\mathcal Q_A$, where $\mathcal Q_F$ and $\mathcal Q_A$ denote the finding-presence and attribute query sets supported by the module. Patient context $c$ comprises the other two retrieved reports, the patient's age, sex, and indication, and prior reports when available, together with the time since the most recent examination. The draft and contextual reports are supplied with their finding labels. At turn $t$, the policy selects one query conditioned on $r_0$, $c$, and the history of previous queries and tool feedback $\tau_{t-1}$:
\begin{equation}
    s_t=(r_0,c,\tau_{t-1}), \qquad
    q_t\sim\pi_\phi(\cdot\mid s_t),
    \label{eq:query_policy}
\end{equation}
where $\tau_0=\emptyset$ and $\pi_\phi$ denotes the query policy. For a valid query, the perception module returns $o_t=\mathcal T_\theta(x,q_t)$. Invalid, repeated, or over-limit queries return error feedback without adding evidence. The query and its feedback are appended to the history before the next decision. Each episode has a budget of $K$ turns, including invalid attempts, with at most $K_F$ finding and $K_A$ attribute queries.

\paragraph{Correction reward.}
Since a queried answer replaces the corresponding draft content, we reward queries that target draft errors and penalize queries that overturn correct content.
We score each executed query by comparing the draft and perception answers against the reference, separately for finding and attribute queries.
For a finding query $q$, we compare the draft answer and the perception prediction with the same reference label:
\begin{equation}
    v_F(q)=
    \begin{cases}
        w_q, & \text{draft wrong, perception correct},\\
        \alpha w_q, & \text{draft wrong, perception wrong},\\
        -w_q, & \text{draft correct, perception wrong},\\
        0, & \text{otherwise},
    \end{cases}
    \label{eq:finding_query_reward}
\end{equation}
where $w_q=(1/p_q)/\bigl(|\mathcal Q_F|^{-1}\sum_{j\in\mathcal Q_F}1/p_j\bigr)$ prioritizes rare findings, $p_q$ is the finding's training prevalence, and the normalization gives $w_q$ unit mean over $\mathcal Q_F$.
The second case awards partial credit $\alpha w_q$, with $\alpha\in[0,1]$, for targeting a draft error even when the perception answer is incorrect.
For an attribute query $q$, an LLM judge scores the draft and perception answers against the reference answer. Let $\Delta_q$ denote the perception-minus-draft score gain, defined in Appendix~\ref{app:reward_details}. We define
\begin{equation}
    v_A(q)=\alpha\,[\text{draft errs on }q]+\Delta_q,
    \label{eq:attribute_query_reward}
\end{equation}
where the bracket is one if the draft has an error on query $q$, and zero otherwise.
Summing over the distinct valid queries executed in an episode gives $V_F$ and $V_A$, and the terminal reward is
\begin{equation}
    R=\lambda\frac{V_F}{Z_F}+(1-\lambda)\frac{V_A}{Z_A},
    \label{eq:query_reward}
\end{equation}
where $\lambda$ weights finding against attribute acquisition, and $Z_F$ and $Z_A$ scale each component by the positive reward attainable on the case within its query limits.
Each component is thus expressed relative to its attainable positive reward, so $\lambda$ need not compensate for differences in raw scale. Exact normalization and scoring details are provided in Appendix~\ref{app:reward_details}.

\paragraph{Policy optimization.}
We optimize $\pi_\phi$ to maximize expected terminal reward using Group Relative Policy Optimization (GRPO)~\citep{shao2024deepseekmath}. For each case, we sample multiple trajectories and compute advantages from their relative rewards. Only the query policy is updated; the perception module remains frozen.

\subsection{Reporting Agent: Evidence-Guided Editing}
\label{sec:report_editing}

The acquired patient-specific evidence must be incorporated into a coherent report while preserving radiological style and phrasing. We design an evidence-guided editing procedure that uses an advanced LLM's semantic understanding and text generation capabilities to identify and express necessary report corrections. To limit opportunities for hallucinations and unwanted edits, we restrict its action space by requiring it to generate only sentence-level semantic patches.
Let $H_K$ collect the query--answer pairs in the final history $\tau_K$, excluding error feedback. Given $r_0$ and $H_K$, a frozen LLM editor $f_e$ is instructed to propose minimal, evidence-supported edits, $\mathcal E=f_e(r_0,H_K)$. Finding answers guide corrections to conflicting statements and additions of missing positive findings; attribute answers refine supported findings without establishing their presence. Each edit either replaces an indexed sentence or inserts text after it, citing supporting observations in $H_K$; deletion is represented by an empty replacement. Validated operations are applied at the original sentence indices, preserving all unedited text. The editor remains outside the policy-training loop; its prompt and implementation details are provided in Appendix~\ref{app:report_editor_prompt}.

\subsection{Agentic Report Generation}
\label{sec:workflow}

At inference, the trained policy uses the retrieved draft and patient context to acquire visual evidence, which the fixed editor incorporates into the final report. Algorithm~\ref{alg:workflow} details the interaction. Rad-VL-JEPA and the LLM editor remain frozen during policy RL; only $\pi_\phi$ is updated.

\begin{algorithm}[tbp]
    \captionsetup{font=small,labelfont=bf,labelsep=space}
    \caption{Agentic Report Generation with Rad-VL-JEPA}
    \label{alg:workflow}
    \normalsize
    \textbf{Input:} CT volume $x$, patient context $c$, query budgets $K$, $K_F$, and $K_A$.\\
    \textbf{Components:}\\
    \hspace*{1em}Rad-VL-JEPA perception tool $\mathcal T_\theta$.\\
    \hspace*{1em}RL-trained query policy $\pi_\phi$.\\
    \hspace*{1em}LLM editor $f_e$.\\
    \textbf{Output:} Final report $\hat r$.
    \begin{enumerate}[label=\arabic*:, leftmargin=1.8em, labelsep=0.5em, itemsep=1pt, parsep=0pt, topsep=3pt]
        \item Retrieve the highest-scoring report $r_0$ using Eq.~\ref{eq:interface}.
        \item Index $r_0$ sentences; set $\tau_0\leftarrow\emptyset$.
        \item \textbf{for} $t=1,\ldots,K$ \textbf{do}
        \item \hspace*{1em}$s_t\leftarrow(r_0,c,\tau_{t-1})$; $q_t\sim\pi_\phi(\cdot\mid s_t)$.\hfill\raisebox{-2.5pt}{\includegraphics[height=13pt]{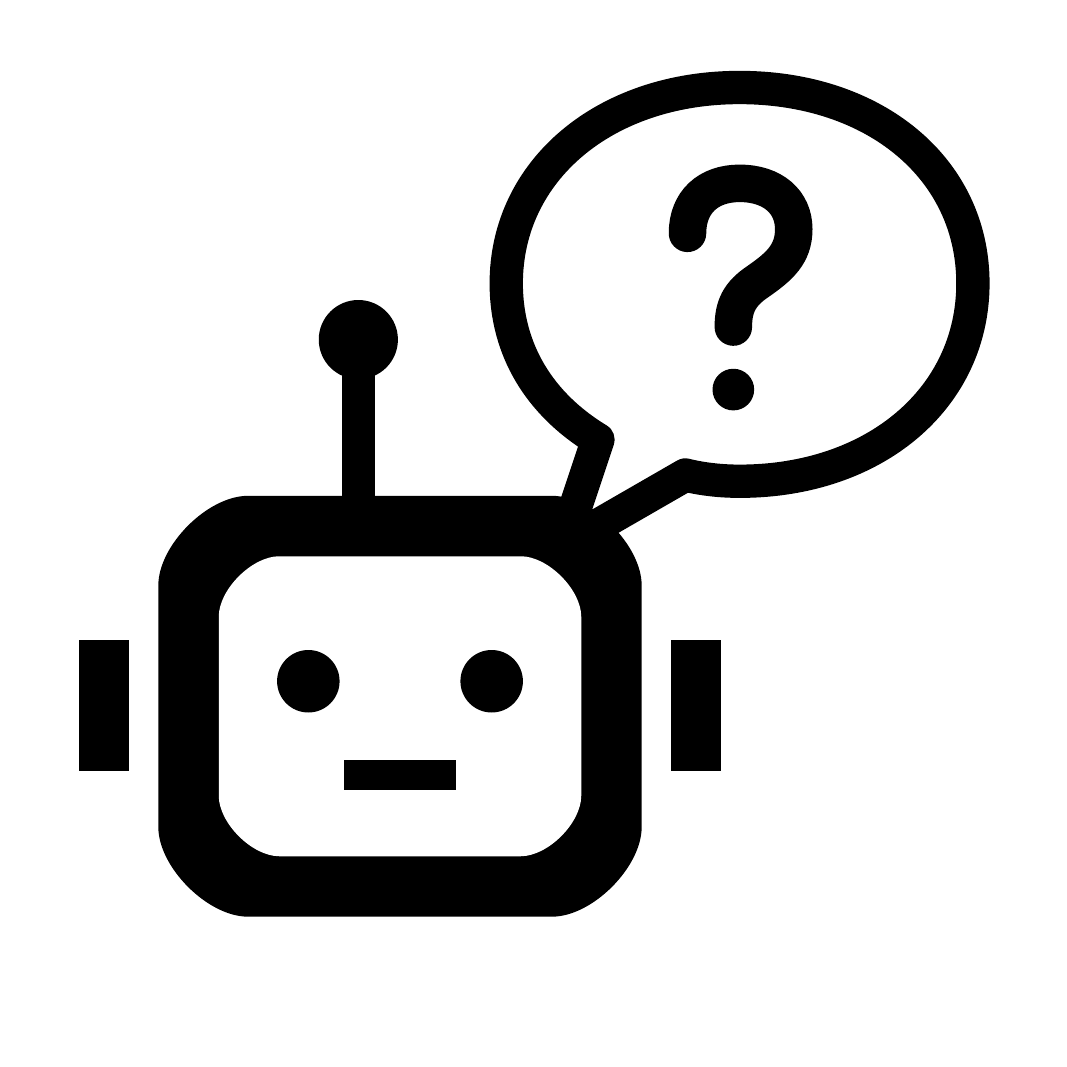}}\;\textcolor{reasoningcolor}{\footnotesize Reasoning}
        \item \hspace*{1em}\parbox[t]{0.67\linewidth}{$o_t\leftarrow\mathcal T_\theta(x,q_t)$, or error feedback if $q_t$ is invalid.}\hfill\raisebox{-2.5pt}{\includegraphics[height=11pt]{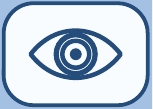}}\;\textcolor{perceptioncolor}{\footnotesize Perception}
        \item \hspace*{1em}Update history $\tau_t$ with $(q_t,o_t)$.
        \item \textbf{end for}
        \item Extract valid query--answer pairs from $\tau_K$ as $H_K$.
        \item $\mathcal E\leftarrow f_e(r_0,H_K)$.\hfill\raisebox{-2.5pt}{\includegraphics[height=13pt]{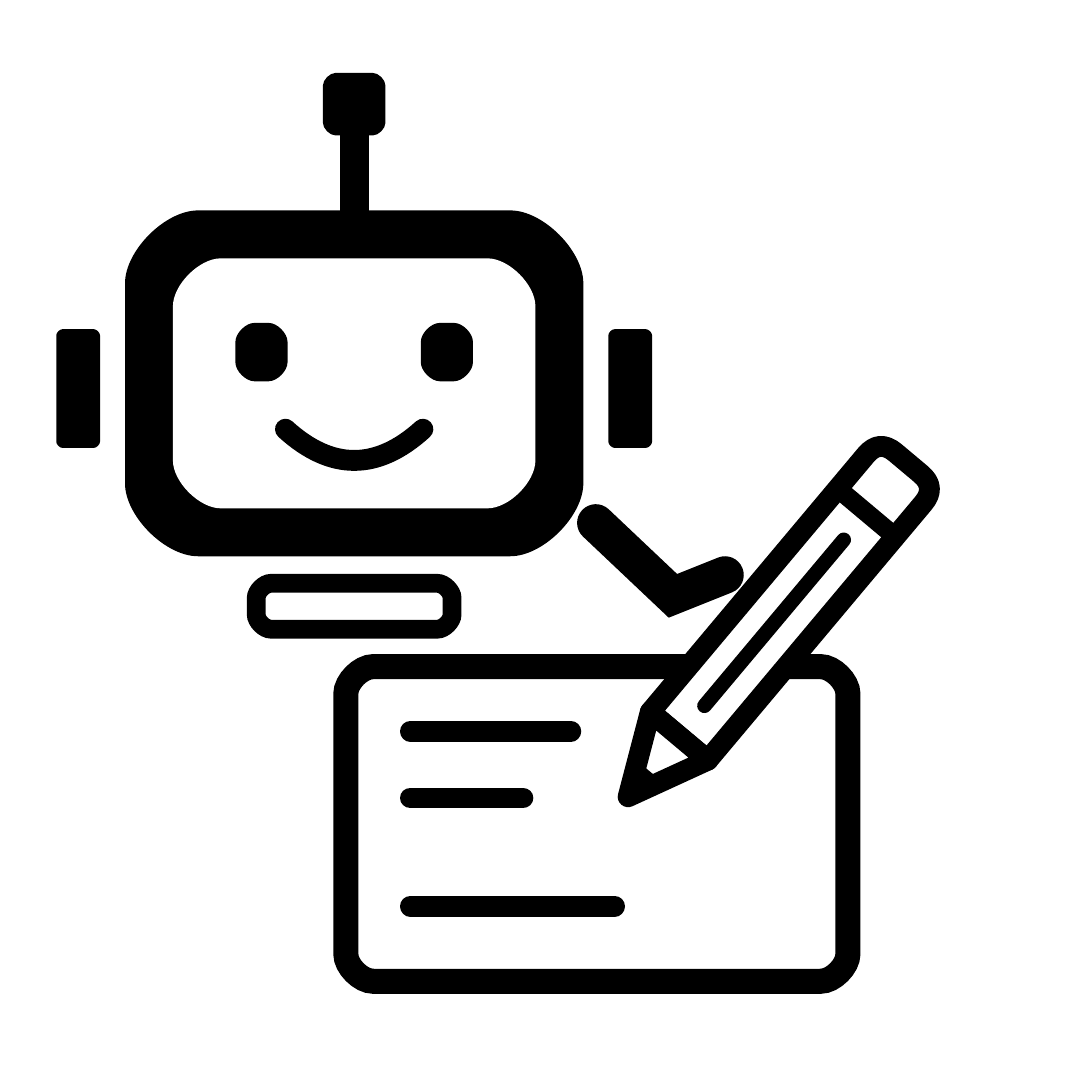}}\;\textcolor{reportingcolor}{\footnotesize Reporting}
        \item \textbf{if} $\mathcal E$ is empty or invalid \textbf{then return} $r_0$.
        \item $\hat r\leftarrow\operatorname{ApplyEdits}(r_0,\mathcal E)$.
        \item \textbf{if} any edit cannot be applied \textbf{then return} $r_0$.
        \item \textbf{return} $\hat r$.
    \end{enumerate}
\end{algorithm}

\section{Experiments and Results}

\begin{table}[tbp]
    \centering
    \captionsetup{font=small,justification=raggedright,singlelinecheck=false}
    \caption{\textbf{Visual perception performance across four tasks (\%).} Higher is better. Metrics are recall at $k$ for zero-shot report-to-CT retrieval, macro AUROC for finding detection, LLM-as-judge score (LLM-J) for attributes, and mean accuracy over seven longitudinal subtasks. Baselines vary by task to match the required retrieval, recognition, or generative capabilities.}
    \label{tab:visual_perception}
    \label{tab:retrieval}
    \label{tab:finding_queries}
    \label{tab:attribute_queries}
    \label{tab:temporal_diagnosis}
    \footnotesize
    \renewcommand{\arraystretch}{1.08}
    \setlength{\tabcolsep}{2pt}
    \begin{tabularx}{\linewidth}{@{}>{\centering\arraybackslash}p{22pt}@{\hspace{17pt}}>{\raggedright\arraybackslash}Xrrr@{}}
    \toprule
    \textbf{Task} & \textbf{Method} & \multicolumn{3}{c}{\textbf{Performance (\%) $\uparrow$}} \\
    \midrule
    \multirow{7.5}{*}{\rotatebox[origin=c]{90}{\shortstack{\textbf{(a) Zero-shot retrieval}\\CT-RATE}}} & & \textbf{R@5} & \textbf{R@10} & \textbf{R@50} \\
    \addlinespace[2pt]
     & Random baseline & 0.3 & 0.6 & 3.2 \\
     & Merlin$^{\dagger}$~\citep{blankemeier2026merlin} & 1.5 & 2.7 & 7.7 \\
     & CT-CLIP~\citep{hamamci2026generalist} & 2.9 & 5.0 & 18.0 \\
     & BrgSA v2~\citep{lai2025brgsa} & 15.0 & 22.2 & 44.4 \\
     & MPS-CT~\citep{li2025performant} & 10.0 & 16.3 & 39.0 \\
    \addlinespace[2pt]
     & \textbf{Rad-VL-JEPA (Ours)} & \textbf{16.3} & \textbf{24.4} & \textbf{48.6} \\
    \midrule
    \multirow{8}{*}{\rotatebox[origin=c]{90}{\shortstack{\textbf{(b) Finding detection}\\CT-RATE}}} & & \multicolumn{3}{c}{\textbf{Macro AUROC}} \\
    \addlinespace[2pt]
     & BIUD$^{\dagger}$~\citep{cao2024biud} & \multicolumn{3}{c}{71.3} \\
     & Merlin$^{\dagger}$~\citep{blankemeier2026merlin} & \multicolumn{3}{c}{72.8} \\
     & CT-CLIP VocabFine~\citep{hamamci2026generalist} & \multicolumn{3}{c}{75.0} \\
     & fVLM~\citep{shui2025fvlm} & \multicolumn{3}{c}{77.8} \\
     & COLIPRI-CM~\citep{wald2026colipri} & \multicolumn{3}{c}{79.8} \\
     & COLIPRI-CRM~\citep{wald2026colipri} & \multicolumn{3}{c}{78.5} \\
    \addlinespace[2pt]
     & \textbf{Rad-VL-JEPA (Ours)} & \multicolumn{3}{c}{\textbf{85.9}} \\
    \midrule
    \multirow{5}{*}{\rotatebox[origin=c]{90}{\shortstack{\textbf{(c) Attributes}\\CT-FineBench}}} & & \multicolumn{3}{c}{\shortstack{\textbf{LLM-as-judge}\\\textbf{score (LLM-J)}}} \\
    \addlinespace[2pt]
     & CT-CHAT 8B~\citep{hamamci2026generalist} & \multicolumn{3}{c}{52.50} \\
     & OmniCT 7B~\citep{lin2026omnict} & \multicolumn{3}{c}{39.13} \\
     & MedM-VL-CT 3B~\citep{shi2025medmvl} & \multicolumn{3}{c}{54.25} \\
    \addlinespace[2pt]
     & \textbf{Rad-VL-JEPA (Ours)} & \multicolumn{3}{c}{\textbf{67.74}} \\
    \midrule
    \multirow{8}{*}{\rotatebox[origin=c]{90}{\shortstack{\textbf{(d) Longitudinal}\\3D-RAD-Bench}}} & & \multicolumn{3}{c}{\textbf{Mean accuracy}} \\
    \addlinespace[2pt]
     & RadFM 13B$^{\dagger}$~\citep{wu2025radfm} & \multicolumn{3}{c}{42.99} \\
     & M3D (Phi3-4B)$^{\dagger}$~\citep{bai2024m3d} & \multicolumn{3}{c}{24.31} \\
     & OmniV (Qwen2.5-1.5B)$^{\dagger}$~\citep{jiang2025omnivmed} & \multicolumn{3}{c}{24.23} \\
     & M3D-RAD (Llama2-7B)~\citep{gai2025threedrad} & \multicolumn{3}{c}{74.78} \\
     & Photon 3B~\citep{fang2026photon} & \multicolumn{3}{c}{77.01} \\
     & Photon Max~\citep{fang2026photon} & \multicolumn{3}{c}{79.70} \\
    \addlinespace[2pt]
     & \textbf{Rad-VL-JEPA (Ours)} & \multicolumn{3}{c}{\textbf{80.94}} \\
    \bottomrule
    \end{tabularx}
    \par\vspace{4pt}
    {\raggedright\footnotesize $^{\dagger}$No component trained on CT-RATE or derived data. Baseline protocols and class sets: Appendix~\ref{app:perception_evaluation}. Longitudinal results follow 3D-RAD-Bench~\citep{gai2025threedrad}.\par}
\end{table}

\subsection{Experimental setup}
\label{sec:setup}
\paragraph{Implementation.}
\textcolor{perceptioncolor}{\textbf{Perception:}} The visual encoder is a frozen Pillar-0~\citep{agrawal2025pillar} chest CT encoder. The target encoder is a frozen Qwen3-Embedding-8B~\citep{zhang2025qwen3}. The predictor is a bidirectional Llama-3.2-1B, and its trainable layers are optimized with AdamW. Training data comprise CT-RATE~\citep{hamamci2026generalist} report sections (impressions and findings) and finding labels; RadGenome-ChestCT~\citep{zhang2025development} region reports, finding annotations, and size QA; ReXGroundingCT finding lists; CT-FineBench~\citep{yuan2026ct} attributes; MeasureVQA~\mbox{\citep{liang2026cheap}} measurement QA; and 3D-RAD~\citep{gai2025threedrad} Tasks~1/2, 3, and 6 for finding recognition and characterization, medical computation, and longitudinal reasoning, respectively. We train the combined VQA corpus for two epochs.
\textcolor{reasoningcolor}{\textbf{Reasoning:}} For evidence acquisition, we evaluate Qwen3 policies from 0.6B to 8B with 24 interaction turns and limits of 12 finding and 12 attribute queries. The report-generation configuration uses the Qwen3-1.7B RL policy with the fixed report editor. The general agent prompt is provided in Appendix~\ref{app:query_plan_prompt_text}.
\textcolor{reportingcolor}{\textbf{Reporting:}} We instantiate the report editor with Claude Opus 5.5 at high reasoning effort; its prompt and implementation details are provided in Appendix~\ref{app:report_editor_prompt}.
\begin{figure*}[!t]
    \centering
    \includegraphics[width=\linewidth]{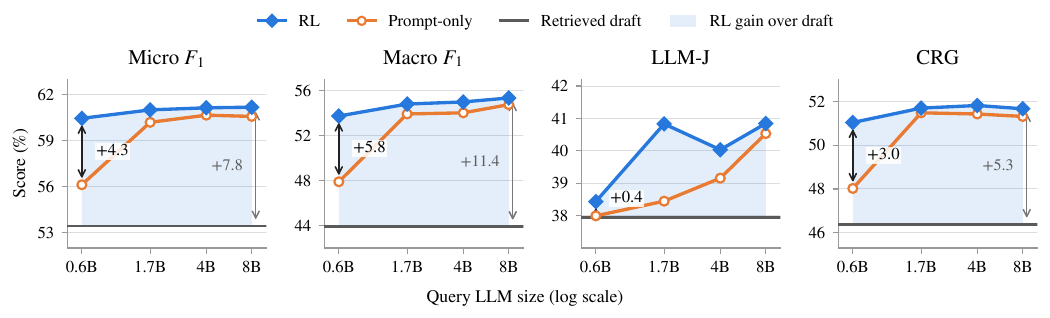}
    \caption{\textbf{RL improves evidence acquisition before report editing.} Scores are percentages; LLM-J is volume-averaged. Blue and orange denote RL and Prompt-only; the gray line is the retrieved draft. Shading spans the draft and RL scores. Black arrows compare RL with Prompt-only; gray arrows compare RL with the draft. Arrow values are percentage-point differences.}
    \label{fig:rl_evidence}
\end{figure*}
\begin{figure*}[!t]
    \centering
    \includegraphics[width=\linewidth]{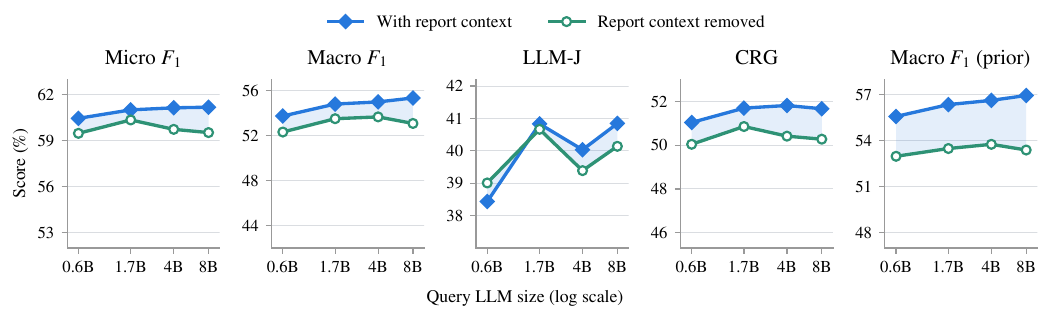}
    \caption{\textbf{Report context improves finding acquisition.} For each policy size, we evaluate the same RL checkpoint with (blue) and without (green) report context. Report context consists of two additional retrieved reports, available prior reports, their finding labels, and the prior-visit interval. Both conditions retain the draft and its finding labels, age, sex, and indication. Scores are percentages; LLM-J is volume-averaged. The last panel evaluates only cases with a prior report.}
    \label{fig:context_gain}
\end{figure*}

\begin{table*}[!t]
\caption{\textbf{Chest CT report generation on CT-RATE.} Scores are fractions; higher is better. P and R are macro-averaged precision and recall; LLM-J is the study-averaged CT-FineBench attribute score. B-1, B-4, and MET denote BLEU-1, BLEU-4, and METEOR. $\dagger$ denotes no training on CT-RATE or derived data; $\ddagger$ marks estimates from published RadAgent results (Appendix~\ref{app:report_evaluation}). -- denotes an unreported metric.}
    \label{tab:reports}
    \centering
    \scriptsize
    \setlength{\tabcolsep}{3.2pt}
    \resizebox{\linewidth}{!}{%
    \begin{tabular}{@{}lccccccccc@{}}
        \toprule
        & \multicolumn{4}{c}{\textbf{Finding presence}} & \textbf{CT-FineBench} & \textbf{CRG} & \multicolumn{3}{c}{\textbf{NLG}} \\
        \cmidrule(lr){2-5}\cmidrule(lr){6-6}\cmidrule(lr){7-7}\cmidrule(lr){8-10}
        \textbf{Method} & \shortstack{\textbf{Micro}\\$\boldsymbol{F_1}$} & \shortstack{\textbf{Macro}\\$\boldsymbol{F_1}$} & \textbf{P} & \textbf{R} & \textbf{LLM-J} & & \textbf{B-1} & \textbf{B-4} & \textbf{MET} \\
        \midrule
        \multicolumn{10}{@{}l}{\textit{VLM-based}} \\
        MedGemma-1.5-4B-IT$^{\dagger}$~\citep{sellergren2026medgemma} & 0.221 & 0.192 & 0.451 & 0.156 & 0.096 & 0.359 & 0.140 & 0.009 & 0.051 \\
        CT-CHAT 8B~\citep{hamamci2026generalist} & 0.294 & 0.174 & 0.332 & 0.176 & 0.194 & 0.369 & 0.404 & 0.208 & 0.198 \\
        BTB3D 8B~\citep{hamamci2025better} & 0.265 & 0.210 & 0.208 & 0.230 & 0.214 & 0.352 & 0.376 & 0.136 & 0.176 \\
        ClinFusion-8B~\citep{yuan2026clinfusion} & 0.297 & 0.215 & 0.410 & 0.162 & 0.200 & 0.373 & 0.371 & 0.159 & 0.178 \\
        Hulu-Med-7B~\citep{jiang2025hulu} & 0.287 & 0.236 & 0.363 & 0.183 & 0.165 & 0.370 & 0.139 & 0.055 & 0.102 \\
        OmniCT 7B~\citep{lin2026omnict} & 0.297 & 0.240 & 0.305 & 0.209 & 0.184 & 0.369 & 0.340 & 0.137 & 0.155 \\
        HSENet 4B~\citep{shi2025hsenet} & 0.387 & 0.304 & 0.458 & 0.251 & 0.275 & 0.397 & 0.392 & 0.212 & 0.200 \\
        Hulu-Med-32B~\citep{jiang2025hulu} & 0.372 & 0.311 & 0.425 & 0.267 & 0.252 & 0.392 & 0.400 & 0.190 & 0.195 \\
        Ker-VLJEPA 3B~\citep{bumgardner2026curriculum} & 0.377 & 0.331 & 0.243 & \textbf{0.597} & 0.386 & 0.391 & 0.330 & 0.071 & 0.150 \\
        \midrule
        \multicolumn{10}{@{}l}{\textit{Agent}} \\
        RadAgent 14B (RL)$^{\ddagger}$~\citep{roschewitz2026radagent} & 0.325 & 0.222 & -- & -- & -- & -- & -- & -- & -- \\
        \textbf{Rad-VL-JEPA Agent 1.7B (Ours)} & \textbf{0.596} & \textbf{0.535} & \textbf{0.532} & 0.557 & \textbf{0.403} & \textbf{0.507} & \textbf{0.500} & \textbf{0.257} & \textbf{0.238} \\
        \bottomrule
    \end{tabular}}
\end{table*}
\begin{figure*}[!t]
    \centering
    \includegraphics[width=0.9\linewidth]{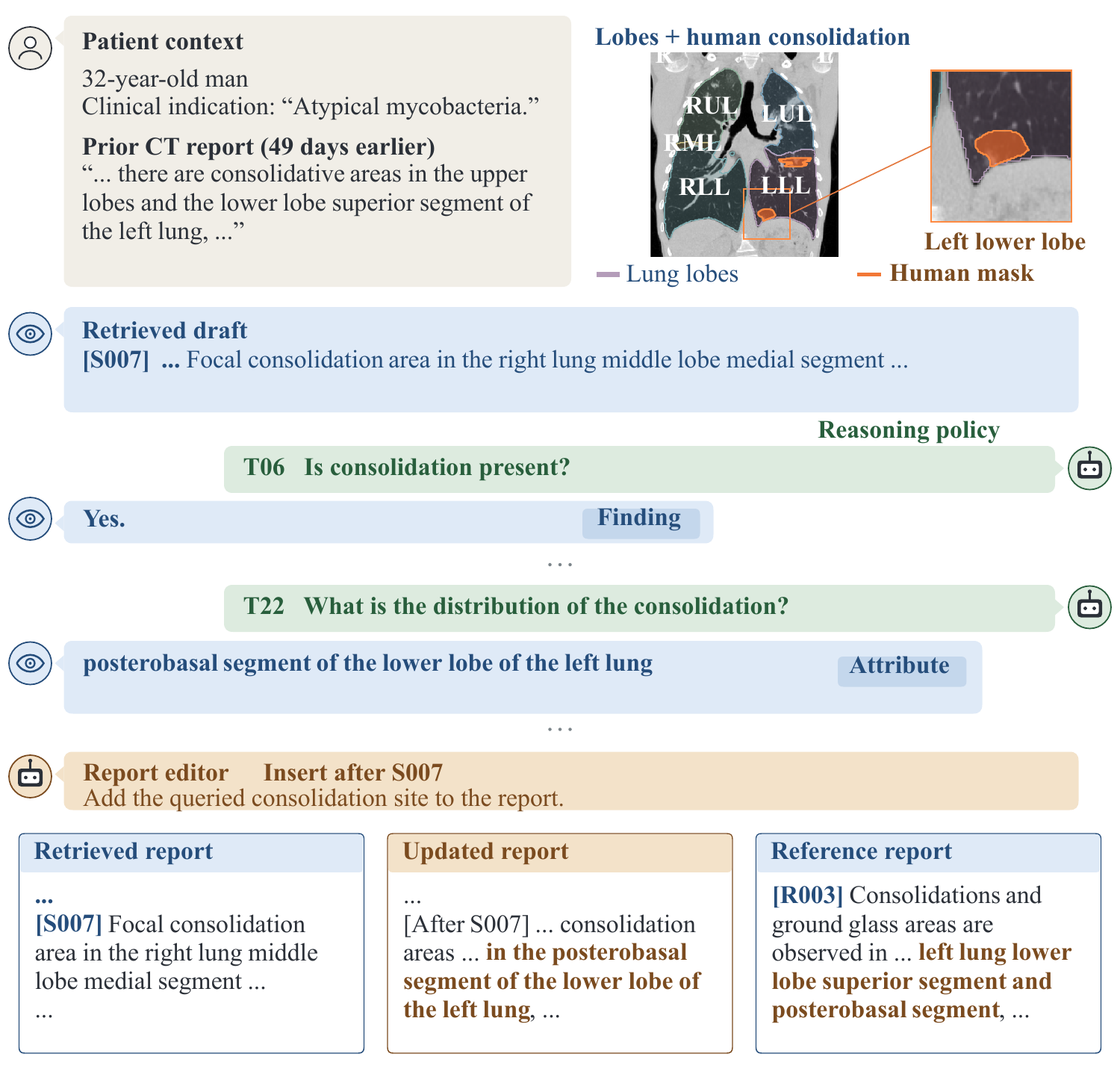}
    \caption{\textbf{Spatial attribute refinement in a real evaluated case.} Given the draft and patient context, $\pi_\phi$ confirms consolidation and queries its distribution. Rad-VL-JEPA returns the left lower-lobe posterobasal segment, which the editor inserts after S007. The overlays use the human consolidation mask from ReXGroundingCT~\citep{baharoon2025rexgroundingct} and TotalSegmentator lung-lobe masks from CT-RATE~\citep{hamamci2026generalist}, for illustration only. Brown marks matching spatial information. Ellipses omit turns and report text. Appendix~\ref{app:qualitative_workflow} retains complete reports and additional cases.}
    \label{fig:case_summary}
\end{figure*}

\paragraph{Evaluation.}
We follow the official CT-RATE validation split for retrieval, finding detection, and report generation; the CT-FineBench protocol for attribute characterization; and the 3D-RAD-Bench test protocol for longitudinal diagnosis. For CT understanding, we report recall at $k$ for report-to-CT retrieval, macro AUROC across 18 findings, CT-FineBench LLM-as-judge scores (LLM-J), and mean accuracy across the seven longitudinal subtasks. Report quality is assessed using finding micro and macro $F_1$, macro precision and recall, CRG score~\citep{hamamci2025crg}, CT-FineBench attribute scores, and BLEU-1, BLEU-4, and METEOR. In Tables~\ref{tab:visual_perception} and~\ref{tab:reports}, $\dagger$ marks methods with no component trained on CT-RATE or derived data. Detailed evaluation protocols are provided in Appendix~\ref{app:evaluation_details}. Evidence-acquisition scores are measured before report editing. Complete scores and paired patient-bootstrap intervals are provided in Appendix~\ref{app:evidence_acquisition_numbers}.
\subsection{Unified Perception Supports Diverse CT Tasks}

\paragraph{Zero-shot report-to-CT retrieval.}
Rad-VL-JEPA uses fixed report-level questions to predict an embedding for each CT, then ranks candidates by similarity to the query report's combined impression and findings embedding. As shown in Table~\ref{tab:retrieval}(a), our model exceeds CT-CLIP, Merlin, MPS-CT, and BrgSA v2~\citep{lai2025brgsa} across all retrieval cutoffs. These results show that the shared embedding space supports report-level retrieval alongside finding and attribute queries.
The improvement over directly using our frozen visual backbone's representations under the same protocol (Appendix~\ref{app:retrieval_ablation}) further indicates benefits beyond the pretrained backbone alone.

\paragraph{Finding detection.}
Rad-VL-JEPA achieves the highest macro AUROC among the listed methods on CT-RATE (Table~\ref{tab:finding_queries}(b)), with a clear margin over fVLM~\citep{shui2025fvlm} and CT-CLIP VocabFine~\citep{hamamci2026generalist}. This suggests that our learned joint embeddings capture discriminative visual features for recognizing clinical findings.

\paragraph{Attribute characterization.}
Rad-VL-JEPA outperforms all evaluated generative VLM baselines on CT-FineBench~\citep{hamamci2026generalist,lin2026omnict,shi2025medmvl}, with a substantial margin over the strongest baseline, MedM-VL-CT (Table~\ref{tab:attribute_queries}(c)). This suggests that our learned joint embeddings capture fine-grained visual details needed to characterize clinical findings.

\paragraph{Longitudinal temporal diagnosis.}
Rad-VL-JEPA achieves the highest mean accuracy among the listed methods on 3D-RAD-Bench, including task-finetuned Photon and Photon Max~\citep{fang2026photon} (Table~\ref{tab:temporal_diagnosis}(d)). This suggests that the joint embeddings support reasoning about finding changes with historical context.

\paragraph{Efficiency.}
We compare Rad-VL-JEPA with a generative VLM under matched visual backbones, training data, and a one-epoch schedule. Figure~\ref{fig:attribute_efficiency} in Appendix~\ref{app:efficiency_evaluation} shows higher best@1 attribute scores and comparable best@3 oracle performance, with fewer trainable parameters and approximately 70\% lower training cost. With visual tokens cached, batched best@1 querying is 5.3$\times$ faster than the generative VLM baseline trained with
the same CT encoder and QA data. Appendix~\ref{app:efficiency_evaluation} gives the training and timing protocols.

\subsection{RL Improves Evidence Acquisition}
\label{sec:query_selection_results}
Figure~\ref{fig:rl_evidence} shows the largest RL gains for the 0.6B policy, whose finding micro $F_1$ nearly matches the 8B Prompt-only policy. Prompt-only finding scores largely plateau across the larger models, which gain less from RL. With the visual tool fixed, this pattern suggests that RL improves how smaller policies select queries, reducing the need to scale the policy for finding acquisition.
Figure~\ref{fig:context_gain} shows a complementary benefit from report context. Adding context improves finding macro $F_1$ across model sizes, with larger gains on cases that have a prior report. This pattern is consistent with contextual reports helping the policy focus its limited query budget on findings relevant to the current scan.
Attribute gains are less uniform: the largest RL gain occurs at 1.7B, while adding report context slightly lowers the 0.6B score. Variation in attribute wording and specificity, together with LLM-based scoring, may contribute to this behavior.
\subsection{Evidence-Guided Edits Improve Reporting}

Table~\ref{tab:reports} shows that our agent achieves the highest macro precision, finding micro $F_1$, and finding macro $F_1$ among the listed methods. The matched draft--editor comparison (Table~\ref{tab:report_editor_results} in Appendix~\ref{app:report_editing_results}) also shows improvements from evidence-guided editing. Patient context guides queries toward relevant findings, Rad-VL-JEPA supplies visual answers, and the editor uses this evidence to correct unsupported claims and add missing findings. Minimal in-place edits retain much of the retrieved draft's wording and structure while updating its clinical content, yielding strong BLEU and METEOR scores alongside improved clinical metrics. This decoupled design allows a 1.7B query policy, paired with a frozen perception tool and a separate report editor, to outperform larger VLM baselines without requiring the policy itself to perform visual interpretation and report writing.

\subsection{Case Study}

Figure~\ref{fig:case_summary} illustrates how context, query selection, and report editing combine in a spatially specific update. The patient's indication is atypical mycobacteria, and a prior report describes consolidation involving the left lower lobe. The retrieved draft provides a structured starting point but describes consolidation in the right middle lobe, omitting the left lower-lobe site. The reasoning agent confirms consolidation and then requests its distribution. Rad-VL-JEPA returns the left lower-lobe posterobasal segment, matching the same finding and site in the reference and human annotation. The editor inserts this site after S007. The trace links contextual input, fine-grained perception, and targeted editing: the attribute answer supplies spatial information beyond a presence-only answer. The CT overlays lung-lobe regions and the human consolidation mask; both are withheld from the agents. Appendix~\ref{app:qualitative_workflow} provides complete reports and additional examples.

\section{Conclusions}

We formulate chest CT report generation as coordinated visual question answering, decoupling visual perception from reasoning and report writing. We introduce Rad-VL-JEPA, a non-generative perception model that unifies report retrieval, finding recognition, and fine-grained attribute characterization through joint-embedding prediction. An RL-trained agent selects patient-specific visual queries, and a separate editor incorporates the acquired evidence through targeted report revisions. Experiments demonstrate strong fine-grained CT understanding with lower training cost and faster inference than a matched generative VLM. RL improves evidence acquisition and helps the 0.6B RL query policy narrow the finding-performance gap to larger prompted policies.


{\small
    \bibliographystyle{ieeenat_fullname}
    \bibliography{references}

@inproceedings{fang2026photon,
  title={Photon: Speedup Volume Understanding with Efficient Multimodal Large Language Models},
  author={Fang, Chengyu and Guo, Heng and Jiang, Zheng and He, Chunming and Li, Xiu and Xu, Minfeng},
  booktitle={International Conference on Learning Representations},
  year={2026},
  url={https://proceedings.iclr.cc/paper_files/paper/2026/hash/633780c1344d0c95e4d2dd3431fe08d9-Abstract-Conference.html}
}

@inproceedings{wald2026colipri,
  title={Comprehensive Language--Image Pre-training for 3D Medical Image Understanding},
  author={Wald, Tassilo and Hamamci, Ibrahim Ethem and Gao, Yuan and Bond-Taylor, Sam and Sharma, Harshita and Ilse, Maximilian and Lo, Cynthia and Melnichenko, Olesya and Schwaighofer, Anton and Codella, Noel C. F. and Wetscherek, Maria Teodora and Maier-Hein, Klaus H. and Korfiatis, Panagiotis and Salvatelli, Valentina and Alvarez-Valle, Javier and P{\'e}rez-Garc{\'i}a, Fernando},
  booktitle={European Conference on Computer Vision},
  year={2026},
  pages={388--406},
  doi={10.1007/978-3-032-37624-4_24},
  url={https://arxiv.org/html/2510.15042v3}
}

@inproceedings{liu2026mimicking,
  title={Mimicking Radiologists: A Coarse-to-Fine Framework with Structural Sparse Tokens for Dual-LLM Computed Tomography Report Generation},
  author={Liu, Hong and Wei, Dong and Zheng, Yefeng and Wu, Xian and Wang, Liansheng},
  booktitle={European Conference on Computer Vision},
  year={2026},
  url={https://media.eventhosts.cc/Conferences/ECCV2026/pdfs/9339.pdf}
}

@article{hamamci2026generalist,
  title={Generalist foundation models from a multimodal dataset for 3D computed tomography},
  author={Hamamci, Ibrahim Ethem and Er, Sezgin and Wang, Chenyu and Almas, Furkan and Simsek, Ayse Gulnihan and Esirgun, Sevval Nil and Dogan, Irem and Durugol, Omer Faruk and Hou, Benjamin and Shit, Suprosanna and others},
  journal={Nature Biomedical Engineering},
volume={10},
pages={1610--1628},
year={2026},
doi={10.1038/s41551-025-01599-y},
  publisher={Nature Publishing Group UK London}
}

@article{hamamci2025better,
  title={Better tokens for better 3d: Advancing vision-language modeling in 3d medical imaging},
  author={Hamamci, Ibrahim Ethem and Er, Sezgin and Shit, Suprosanna and Reynaud, Hadrien and Yang, Dong and Guo, Pengfei and Edgar, Marc and Xu, Daguang and Kainz, Bernhard and Menze, Bjoern},
  journal={Advances in Neural Information Processing Systems},
  volume={38},
  pages={135074--135102},
  year={2025}
}

@misc{shi2025hsenet,
  title={HSENet: Hybrid Spatial Encoding Network for 3D Medical Vision-Language Understanding}, 
      author={Yanzhao Shi and Xiaodan Zhang and Junzhong Ji and Haoning Jiang and Chengxin Zheng and Yinong Wang and Liangqiong Qu},
      year={2025},
      eprint={2506.09634},
      archivePrefix={arXiv},
      primaryClass={cs.CV},
      url={https://arxiv.org/abs/2506.09634}, 
}

@inproceedings{lin2026omnict,
  title={{OmniCT}: Towards a Unified Slice-Volume {LVLM} for Comprehensive {CT} Analysis},
  author={Lin, Tianwei and Qiu, Zhongwei and Zhang, Wenqiao and Liu, Jiang and Xie, Yihan and Gao, Mingjian and Fan, Zhenxuan and Li, Zhaocheng and Li, Sijing and Xie, Zhongle and others},
  booktitle={International Conference on Learning Representations},
  year={2026}
}

@inproceedings{shi2025medmvl,
author="Shi, Yiming
and Yang, Shaoshuai
and Zhu, Xun
and Wang, Haoyu
and Fu, Xiangling
and Li, Miao
and Wu, Ji",
title="MedM-VL: What Makes a Good Medical LVLM?",
booktitle="AI for Clinical Applications",
year="2026",
publisher="Springer Nature Switzerland",
address="Cham",
pages="290--299",
isbn="978-3-032-06004-4"
}

@inproceedings{li2025performant,
  title={More Performant and Scalable: Rethinking Contrastive Vision-Language Pre-training of Radiology in the {LLM} Era},
  author={Li, Yingtai and Lai, Haoran and Zhou, Xiaoqian and Ming, Shuai and Ma, Wenxin and Wei, Wei and Zhou, S. Kevin},
  booktitle={Medical Image Computing and Computer Assisted Intervention -- MICCAI 2025},
  year={2025},
  publisher={Springer Nature Switzerland}
}

@article{lai2025brgsa,
  author={Lai, Haoran and Jiang, Zihang and Yao, Qingsong and Wang, Rongsheng and He, Zhiyang and Tao, Xiaodong and Lv, Weifu and Wei, Wei and Zhou, Shaohua Kevin},
  journal={IEEE Journal of Biomedical and Health Informatics}, 
  title={Bridged Semantic Alignment for Zero-Shot 3D Medical Image Diagnosis}, 
  year={2025},
  volume={},
  number={},
  pages={1-14},
  doi={10.1109/JBHI.2025.3629096}}

@article{blankemeier2026merlin,
  title={Merlin: a computed tomography vision--language foundation model and dataset},
  author={Blankemeier, Louis and Kumar, Ashwin and Cohen, Joseph Paul and Liu, Jiaming and Liu, Longchao and Van Veen, Dave and Gardezi, Syed Jamal Safdar and Yu, Hongkun and Paschali, Magdalini and Chen, Zhihong and others},
  journal={Nature},
  volume={652},
  number={8112},
  pages={1318--1328},
  year={2026},
  publisher={Nature Publishing Group UK London}
}

@misc{roschewitz2026radagent,
      title={RadAgent: A tool-using AI agent for stepwise interpretation of chest computed tomography}, 
      author={Mélanie Roschewitz and Kenneth Styppa and Yitian Tao and Jiwoong Sohn and Jean-Benoit Delbrouck and Benjamin Gundersen and Nicolas Deperrois and Christian Bluethgen and Julia E. Vogt and Bjoern Menze and Farhad Nooralahzadeh and Michael Krauthammer and Michael Moor},
      year={2026},
      eprint={2604.15231},
      archivePrefix={arXiv},
      primaryClass={cs.AI},
      url={https://arxiv.org/abs/2604.15231}, 
}

@inproceedings{shui2025fvlm,
  author       = {Zhongyi Shui and
                  Jianpeng Zhang and
                  Weiwei Cao and
                  Sinuo Wang and
                  Ruizhe Guo and
                  Le Lu and
                  Lin Yang and
                  Xianghua Ye and
                  Tingbo Liang and
                  Qi Zhang and
                  Ling Zhang},
  title        = {Large-scale and Fine-grained Vision-language Pre-training for Enhanced
                  {CT} Image Understanding},
  booktitle    = {The Thirteenth International Conference on Learning Representations,
                  {ICLR} 2025, Singapore, April 24-28, 2025},
  publisher    = {OpenReview.net},
  year         = {2025},
  url          = {https://openreview.net/forum?id=nYpPAT4L3D},
  bibsource    = {dblp computer science bibliography, https://dblp.org}
}

@inproceedings{cao2024biud,
  title={Bootstrapping Chest {CT} Image Understanding by Distilling Knowledge from X-ray Expert Models},
  author={Cao, Weiwei and Zhang, Jianpeng and Xia, Yingda and Mok, Tony C. W. and Li, Zi and Ye, Xianghua and Lu, Le and Zheng, Jian and Tang, Yuxing and Zhang, Ling},
  booktitle={Proceedings of the IEEE/CVF Conference on Computer Vision and Pattern Recognition},
  pages={11238--11247},
  year={2024}
}

@article{mao2026ct,
  title={Ct-agent: A multimodal-llm agent for 3d ct radiology question answering},
  author={Mao, Yuren and Xu, Wenyi and Qin, Yuyang and Gao, Yunjun},
  journal={Science China Information Sciences},
  volume={69},
  number={5},
  pages={150107},
  year={2026},
  publisher={Springer}
}

@inproceedings{lin2026march,
    title = "{MARCH}: Multi-Agent Radiology Clinical Hierarchy for {CT} Report Generation",
    author = "Lin, Yi  and
      Ding, Yihao  and
      Wu, Yonghui  and
      Peng, Yifan",
    editor = "Liakata, Maria  and
      Moreira, Viviane P.  and
      Zhang, Jiajun  and
      Jurgens, David",
    booktitle = "Proceedings of the 64th Annual Meeting of the {A}ssociation for {C}omputational {L}inguistics (Volume 2: Short Papers)",
    month = jul,
    year = "2026",
    address = "San Diego, California, United States",
    publisher = "Association for Computational Linguistics",
    url = "https://aclanthology.org/2026.acl-short.24/",
    doi = "10.18653/v1/2026.acl-short.24",
    pages = "273--285",
    ISBN = "979-8-89176-391-3"
}

@misc{chen2026vl,
      title={VL-JEPA: Joint Embedding Predictive Architecture for Vision-language}, 
      author={Delong Chen and Mustafa Shukor and Theo Moutakanni and Willy Chung and Jade Yu and Tejaswi Kasarla and Yejin Bang and Allen Bolourchi and Yann LeCun and Pascale Fung},
      year={2026},
      eprint={2512.10942},
      archivePrefix={arXiv},
      primaryClass={cs.CV},
      url={https://arxiv.org/abs/2512.10942}, 
}

@inproceedings{cheng2026rethinking,
  title={Rethinking radiology report generation: From narrative flow to topic-guided findings},
  author={Cheng, Sheng and Subramanian, Devika},
  booktitle={The Fourteenth International Conference on Learning Representations},
  year={2026},
  url={https://proceedings.iclr.cc/paper_files/paper/2026/hash/0f904b512f0cb664c71630b2d276348b-Abstract-Conference.html}
}

@misc{yuan2026clinfusion,
      title={ClinFusion: A Vision-Centric Multimodal LLM System for Holistic Medical Understanding}, 
      author={Hangjie Yuan and Yichen Qian and Zhiwei Tang and Xianzhe Xu and Lirong Wu and Sicheng Yang and Jinwang Wang and Pengju Wang and Zhitao Zeng and Yizeng Han and Yan Xing and Shengxuan Luo and Tao Feng and Qing Xie and Weigen Yao and Yi Yang and Zuozhu Liu and Jiasheng Tang and Shaocheng Wang and Jitao Wang and Jiahong Dong and Weihua Chen and Feng Xu and Fan Wang},
      year={2026},
      eprint={2607.24743},
      archivePrefix={arXiv},
      primaryClass={cs.CV},
      url={https://arxiv.org/abs/2607.24743}, 
}

@misc{jiang2025hulu,
  title={Hulu-Med: A Transparent Generalist Model towards Holistic Medical Vision-Language Understanding}, 
      author={Songtao Jiang and Yuan Wang and Sibo Song and Tianxiang Hu and Chenyi Zhou and Bin Pu and Yan Zhang and Zhibo Yang and Yang Feng and Joey Tianyi Zhou and Jin Hao and Zijian Chen and Ruijia Wu and Tao Tang and Junhui Lv and Hongxia Xu and Hongwei Wang and Jun Xiao and Bin Feng and Fudong Zhu and Kenli Li and Weidi Xie and Jimeng Sun and Jian Wu and Zuozhu Liu},
      year={2025},
      eprint={2510.08668},
      archivePrefix={arXiv},
      primaryClass={cs.CV},
      url={https://arxiv.org/abs/2510.08668}, 
}

@misc{bumgardner2026curriculum,
      title={Curriculum-Driven 3D CT Report Generation via Language-Free Visual Grafting and Zone-Constrained Compression}, 
      author={V. K. Cody Bumgardner and Mitchell A. Klusty and Mahmut S. Gokmen and Evan W. Damron},
      year={2026},
      eprint={2603.23308},
      archivePrefix={arXiv},
      primaryClass={cs.CV},
      url={https://arxiv.org/abs/2603.23308}, 
}

@misc{sellergren2026medgemma,
  title={Medgemma 1.5 technical report},
  author={Sellergren, Andrew and Gao, Chufan and Mahvar, Fereshteh and Kohlberger, Timo and Jamil, Fayaz and Traverse, Madeleine and Tono, Alberto and Sadjad, Bashir and Yang, Lin and Lau, Charles and others},
  year={2026},
      eprint={2604.05081},
      archivePrefix={arXiv},
      primaryClass={cs.AI},
      url={https://arxiv.org/abs/2604.05081}, 
}

@misc{agrawal2025pillar,
      title={Pillar-0: A New Frontier for Radiology Foundation Models}, 
      author={Kumar Krishna Agrawal and Longchao Liu and Long Lian and Michael Nercessian and Natalia Harguindeguy and Yufu Wu and Peter Mikhael and Gigin Lin and Lecia V. Sequist and Florian Fintelmann and Trevor Darrell and Yutong Bai and Maggie Chung and Adam Yala},
      year={2025},
      eprint={2511.17803},
      archivePrefix={arXiv},
      primaryClass={cs.CV},
      url={https://arxiv.org/abs/2511.17803}, 
}

@misc{zhang2025qwen3,
  title={Qwen3 Embedding: Advancing Text Embedding and Reranking Through Foundation Models}, 
      author={Yanzhao Zhang and Mingxin Li and Dingkun Long and Xin Zhang and Huan Lin and Baosong Yang and Pengjun Xie and An Yang and Dayiheng Liu and Junyang Lin and Fei Huang and Jingren Zhou},
      year={2025},
      eprint={2506.05176},
      archivePrefix={arXiv},
      primaryClass={cs.CL},
      url={https://arxiv.org/abs/2506.05176}, 
}

@inproceedings{yuan2026ct,
    title = "{CT}-{F}ine{B}ench: A Diagnostic Fidelity Benchmark for Fine-Grained Evaluation of {CT} Report Generation",
    author = "Yuan, Ruifeng  and
      Chang, Wanxing  and
      Cao, Weiwei  and
      Shi, Bowen  and
      Wei, Zhongyu  and
      Zhang, Ling  and
      Zhang, Jianpeng",
    booktitle = "Proceedings of the 64th Annual Meeting of the {A}ssociation for {C}omputational {L}inguistics (Volume 1: Long Papers)",
    month = jul,
    year = "2026",
    address = "San Diego, California, United States",
    publisher = "Association for Computational Linguistics",
    url = "https://aclanthology.org/2026.acl-long.2038/",
    doi = "10.18653/v1/2026.acl-long.2038",
    pages = "44045--44058",
    ISBN = "979-8-89176-390-6",

}

@article{zhang2025development,
  title={Development of a large-scale grounded vision language dataset for chest CT analysis},
  author={Zhang, Xiaoman and Wu, Chaoyi and Zhao, Ziheng and Lei, Jiayu and Tian, Weiwei and Zhang, Ya and Xie, Weidi and Wang, Yanfeng},
  journal={Scientific Data},
  volume={12},
  number={1},
  pages={1636},
  year={2025},
  publisher={Nature Publishing Group UK London}
}

@misc{liang2026cheap,
      title={When Do Cheap Probes Predict Expensive Training? Probing 3D-CT Encoders for Text Generation}, 
      author={Renjie Liang and Zijian Xu},
      year={2026},
      eprint={2607.22771},
      archivePrefix={arXiv},
      primaryClass={cs.CV},
      url={https://arxiv.org/abs/2607.22771}, 
}

@inproceedings{gai2025threedrad,
  title={3D-RAD: A Comprehensive 3D Radiology Med-VQA Dataset with Multi-Temporal Analysis and Diverse Diagnostic Tasks},
  author={Gai, Xiaotang and Liu, Jiaxiang and Li, Yichen and Meng, Zijie and Wu, Jian and Liu, Zuozhu},
  booktitle={Advances in Neural Information Processing Systems},
  volume={38},
  year={2025}
}

@article{shao2024deepseekmath,
  title={{DeepSeekMath}: Pushing the Limits of Mathematical Reasoning in Open Language Models},
  author={Shao, Zhihong and Wang, Peiyi and Zhu, Qihao and Xu, Runxin and Song, Junxiao and Bi, Xiao and Zhang, Haowei and Zhang, Mingchuan and Li, Y. K. and Wu, Y. and Guo, Daya},
  journal={arXiv preprint arXiv:2402.03300},
  year={2024},
  url={https://arxiv.org/abs/2402.03300}
}

@misc{wang2026enhancing,
      title={Enhancing Fine-Grained Spatial Grounding in 3D CT Report Generation via Discriminative Guidance},
      author={Chenyu Wang and Weicheng Dai and Han Liu and Wenchao Li and Kayhan Batmanghelich},
      year={2026},
      eprint={2604.10437},
      archivePrefix={arXiv},
      primaryClass={cs.CV},
      url={https://arxiv.org/abs/2604.10437},
}

@inproceedings{gao2023rarr,
  title={{RARR}: Researching and Revising What Language Models Say, Using Language Models},
  author={Gao, Luyu and Dai, Zhuyun and Pasupat, Panupong and Chen, Anthony and Chaganty, Arun Tejasvi and Fan, Yicheng and Zhao, Vincent and Lao, Ni and Lee, Hongrae and Juan, Da-Cheng and Guu, Kelvin},
  booktitle={Proceedings of the 61st Annual Meeting of the Association for Computational Linguistics (Volume 1: Long Papers)},
  pages={16477--16508},
  year={2023},
  publisher={Association for Computational Linguistics},
  doi={10.18653/v1/2023.acl-long.910},
  url={https://aclanthology.org/2023.acl-long.910/}
}

@inproceedings{huang2024fineradscore,
  title={{FineRadScore}: A Radiology Report Line-by-Line Evaluation Technique Generating Corrections with Severity Scores},
  author={Huang, Alyssa and Banerjee, Oishi and Wu, Kay and Reis, Eduardo Pontes and Rajpurkar, Pranav},
  booktitle={Proceedings of the 9th Machine Learning for Healthcare Conference},
  series={Proceedings of Machine Learning Research},
  volume={252},
  year={2024},
  publisher={PMLR},
  url={https://proceedings.mlr.press/v252/huang24a.html}
}

@misc{myronenko2026nvreasonct,
  title={{NV-Reason-CT}: {3D} Visual Language Model for {CT} Analysis},
  author={Myronenko, Andriy and Yang, Dong and Tang, Yucheng and Turkbey, Baris and Simon, Benjamin and Harmon, Stephanie and Makwana, Rikhil and Aboian, Mariam and Azamat, Sena and Hamamci, Ibrahim Ethem and Er, Sezgin and Menze, Bjoern and Zhou, Zongwei and Li, Wenxuan and Edgar, Marc and He, Yufan and Guo, Pengfei and Xu, Daguang},
  year={2026},
  eprint={2609.27511},
  archivePrefix={arXiv},
  primaryClass={cs.CV},
  url={https://arxiv.org/abs/2609.27511}
}

@article{hamamci2025crg,
  title={{CRG} Score: A Distribution-Aware Clinical Metric for Radiology Report Generation},
  author={Hamamci, Ibrahim Ethem and Er, Sezgin and Shit, Suprosanna and Reynaud, Hadrien and Kainz, Bernhard and Menze, Bjoern},
  journal={arXiv preprint arXiv:2505.17167},
  year={2025},
  url={https://arxiv.org/abs/2505.17167}
}

@article{wu2025radfm,
  title={Towards generalist foundation model for radiology by leveraging web-scale {2D}\&{3D} medical data},
  author={Wu, Chaoyi and Zhang, Xiaoman and Zhang, Ya and Hui, Hui and Wang, Yanfeng and Xie, Weidi},
  journal={Nature Communications},
  volume={16},
  pages={7866},
  year={2025},
  doi={10.1038/s41467-025-62385-7}
}

@misc{bai2024m3d,
  title={{M3D}: Advancing {3D} Medical Image Analysis with Multi-Modal Large Language Models},
  author={Bai, Fan and Du, Yuxin and Huang, Tiejun and Meng, Max Q.-H. and Zhao, Bo},
  year={2024},
  eprint={2404.00578},
  archivePrefix={arXiv},
  primaryClass={cs.CV}
}

@misc{jiang2025omnivmed,
  title={{OmniV-Med}: Scaling Medical Vision-Language Model for Universal Visual Understanding},
  author={Jiang, Songtao and Wang, Yuan and Song, Sibo and Zhang, Yan and Meng, Zijie and Lei, Bohan and Wu, Jian and Sun, Jimeng and Liu, Zuozhu},
  year={2025},
  eprint={2504.14692},
  archivePrefix={arXiv},
  primaryClass={cs.CL}
}

@article{baharoon2025rexgroundingct,
  title={ReXGroundingCT: A 3D Chest CT Dataset for Segmentation of Findings from Free-Text Reports},
  author={Baharoon, Mohammed and Luo, Luyang and Moritz, Michael and Kumar, Abhinav and Kim, Sung Eun and Zhang, Xiaoman and others},
  journal={arXiv preprint arXiv:2507.22030},
  year={2025}
}

@article{ge2025ipcrr,
  title={{IP-CRR}: Information Pursuit for Interpretable Classification of Chest Radiology Reports},
  author={Ge, Yuyan and Chan, Kwan Ho Ryan and Messina, Pablo and Vidal, Ren{\'e}},
  journal={arXiv preprint arXiv:2505.00191},
  year={2025},
  url={https://arxiv.org/abs/2505.00191}
}

@article{wasserthal2023totalsegmentator,
  title={TotalSegmentator: robust segmentation of 104 anatomic structures in CT images},
  author={Wasserthal, Jakob and Breit, Hanns-Christian and Meyer, Manfred T and Pradella, Maurice and Hinck, Daniel and Sauter, Alexander W and Heye, Tobias and Boll, Daniel T and Cyriac, Joshy and Yang, Shan and others},
  journal={Radiology: Artificial Intelligence},
  volume={5},
  number={5},
  pages={e230024},
  year={2023},
  publisher={Radiological Society of North America}
}
}

\onecolumn
\ifdefined\nolinenumbers\nolinenumbers\fi
\raggedbottom
\setcounter{page}{1}
\renewcommand{\thepage}{S\arabic{page}}
\begin{center}
    \Large\textbf{\thetitle}\\[0.5em]
    Supplementary Material
\end{center}

\appendix
\section*{Appendix Contents}
\pdfbookmark[0]{Appendix Contents}{appendix.contents}
\begingroup
\small
\setlength{\parskip}{0pt}
\newcommand{\appendixcontentsentry}[2]{%
    \noindent\hspace*{#1}%
    \hyperref[#2]{\makebox[2.5em][l]{\ref*{#2}}\nameref*{#2}}%
    \dotfill\hyperref[#2]{\pageref*{#2}}\par
}
{\bfseries\appendixcontentsentry{0pt}{app:implementation_details}}
\appendixcontentsentry{1em}{app:jepa_implementation}
\appendixcontentsentry{1em}{app:answer_banks}
\appendixcontentsentry{1em}{app:agent_prompt}
\appendixcontentsentry{1em}{app:report_editor_prompt}
{\bfseries\appendixcontentsentry{0pt}{app:evaluation_details}}
\appendixcontentsentry{1em}{app:perception_evaluation}
\appendixcontentsentry{1em}{app:efficiency_evaluation}
\appendixcontentsentry{1em}{app:query_selection_evaluation}
\appendixcontentsentry{1em}{app:report_evaluation}
{\bfseries\appendixcontentsentry{0pt}{app:additional_quantitative_results}}
\appendixcontentsentry{1em}{app:retrieval_ablation}
\appendixcontentsentry{1em}{app:evidence_acquisition_numbers}
\appendixcontentsentry{1em}{app:report_editing_results}
\appendixcontentsentry{1em}{app:per_finding_comparison}
{\bfseries\appendixcontentsentry{0pt}{app:qualitative_workflow}}
\appendixcontentsentry{1em}{app:qualitative_consolidation_case}
\appendixcontentsentry{1em}{app:qualitative_nodule_case}
\appendixcontentsentry{1em}{app:additional_spatial_examples}
{\bfseries\appendixcontentsentry{0pt}{app:query_plan_prompt}}
\appendixcontentsentry{1em}{app:query_plan_prompt_text}
\appendixcontentsentry{1em}{app:query_catalog}
\endgroup

\section{Implementation Details}
\label{app:implementation_details}

\paragraph{Data ethics.}
This study uses only publicly released, de-identified research datasets and involves no new participant recruitment or clinical data collection. We did not seek new IRB approval or obtain new participant consent for this secondary analysis of de-identified data. Dataset provenance and collection protocols are attributed to the original providers in Section~\ref{sec:setup}.

\subsection{Rad-VL-JEPA Architecture and Training}
\label{app:jepa_implementation}

\begin{figure}[H]
    \centering
    \includegraphics[width=\linewidth]{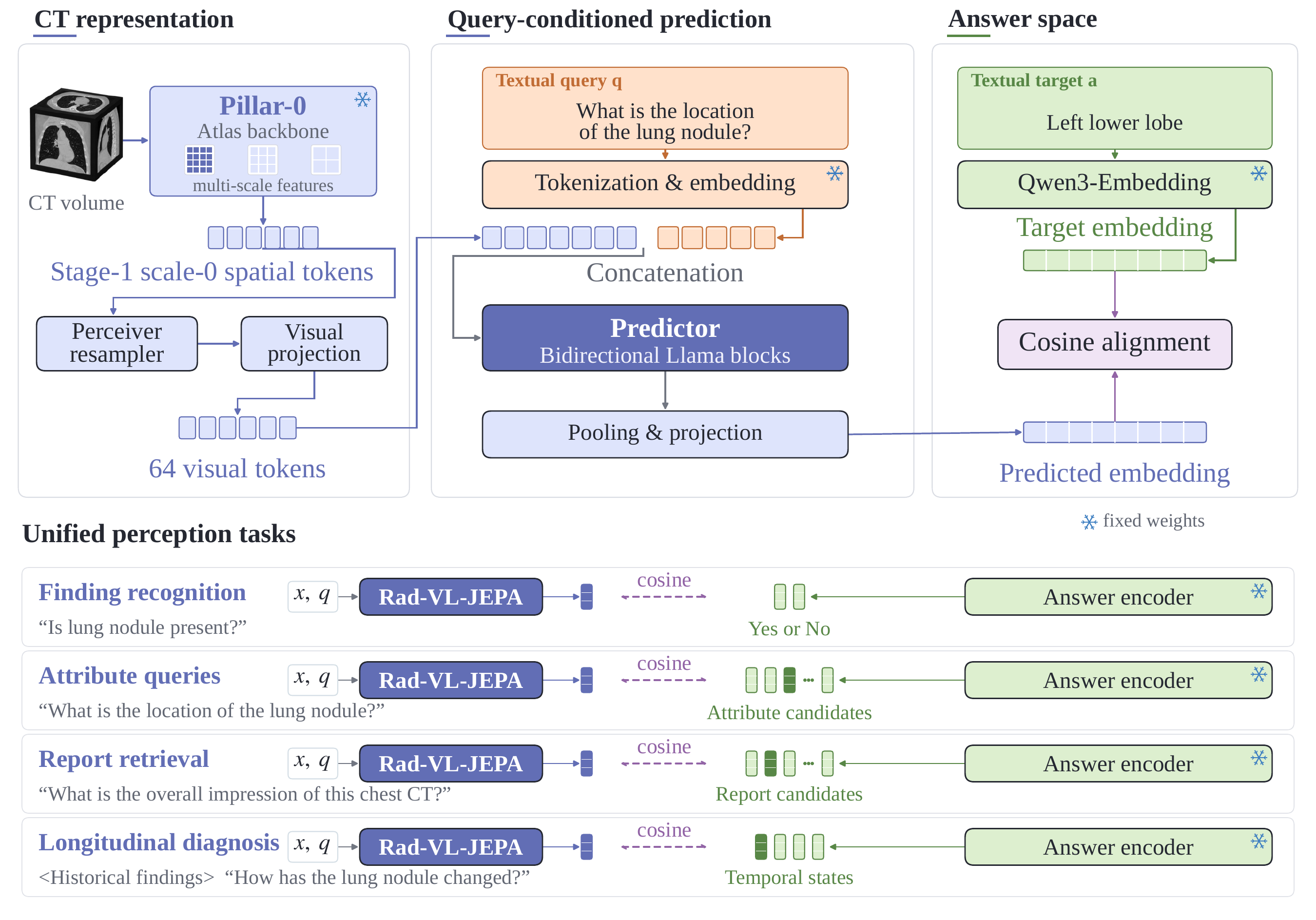}
    \caption{\textbf{Rad-VL-JEPA architecture and tasks.} Top: CT and query embeddings feed a shared predictor, trained by cosine alignment with a fixed text-encoder target. Bottom: query-specific candidate matching supports finding recognition, attribute characterization, report retrieval, and longitudinal diagnosis. Longitudinal queries also incorporate historical finding-presence labels. Colored bars denote embeddings; snowflakes mark fixed encoders or embedding lookups. Candidate scoring follows Section~\ref{sec:perception_agent}.}
    \label{fig:jepa_architecture}
\end{figure}

\paragraph{Architecture.}
Figure~\ref{fig:jepa_architecture} summarizes the modules and their information flow.
The frozen Pillar-0 encoder supplies 1,024 visual tokens of dimension 384. A query-independent Perceiver compresses them into 64 learned latents of dimension 384 using one cross-attention block with six heads and a two-layer GELU feed-forward network of hidden dimension 768. A linear projection maps these latents to the 2,048-dimensional Llama-3.2-1B hidden space. Query tokens use the frozen Llama embedding table. The predictor uses the final eight pretrained Llama layers (indices 8--15), with causal attention replaced by bidirectional attention and padded positions masked. We mean-pool valid visual and query positions, then linearly project to the 4,096-dimensional answer space and normalize. Answer targets are precomputed with frozen Qwen3-Embedding-8B using last-token pooling and normalization; there is no trainable target-side projection.

\paragraph{Optimization and task mixture.}
We train the resampler, visual projection, eight predictor layers, and output projection with the cosine loss in Section~\ref{sec:perception_agent}. The CT encoder, query-token embeddings, and target answer encoder remain frozen. The corpus contains 2,492,454 eligible QA rows spanning the datasets listed in Section~\ref{sec:setup}. Each epoch shuffles the full mixture once, without task reweighting or binary class balancing. Training uses seed 0, two B200 GPUs, a global batch of 512 QA rows, and two epochs (9,736 updates). AdamW uses learning rate $5\times10^{-5}$, betas $(0.9,0.95)$, and 500 warm-up updates; the rate is constant after warm-up through the training endpoint. Weight decay is 0.24 on matrices and zero on biases, normalization parameters, and learned latents. This perception training precedes policy RL. During policy RL and report editing, all perception parameters are frozen.

\clearpage
\subsection{Answer Banks and Finding Calibration}
\label{app:answer_banks}

\paragraph{Answer banks.}
Report retrieval uses training-set reports and excludes reports from the same patient. For the reporting tool's attribute queries, we collect answers from the CT-FineBench training QA, group them by finding and canonical attribute type, and deduplicate exact answer strings. An LLM (GPT-6.1-sol) audits each query's candidate pool and removes candidates judged to be off-topic. We precompute the embeddings of retained candidates using the fixed answer encoder. At inference, each attribute query ranks only its corresponding bank; its answer characterizes a finding conditionally and does not establish finding presence. Binary finding recognition uses the calibrated \textsc{Yes}--\textsc{No} margin described below.

\paragraph{Answer-bank cleaning prompt.}
We use the following instruction with GPT-6.1-sol at medium reasoning effort, supplying one attribute question and its full training-answer pool per request. The input includes the finding, attribute, question, attribute definition, dataset examples, and numbered candidates. The returned flags are checked against the candidate indices and exact answer strings. Only candidates flagged \texttt{I} are removed; \texttt{U} and unlisted candidates are retained. Angle-bracketed fields in the input template are filled for each query.

\noindent\begin{minipage}{\linewidth}
\begin{promptbox}{Answer-bank cleaning: instruction}
You clean an answer bank for chest CT attribute questions. Each candidate is a short answer
phrase taken from radiology reports and filed under one (finding, attribute) question. Find
the candidates that do NOT answer that question (off-topic). You are not judging whether an
answer is clinically correct for any patient.

Keep (do not list) a candidate if ANY part of it states a value of the asked attribute for
this finding, even when it is terse, vague, broad, hedged, misspelled, oddly phrased, or also
mentions other attributes, other findings or extra anatomy. Interpret the attribute through
its definition (e.g. if severity covers extent, 'widespread' answers it).
List as "I" (off-topic) a candidate when NO part of it answers the asked attribute, for
example: it only gives a different attribute (a location for severity, 'millimetric' or a size
for number, a type for severity, a change over time such as 'regressed' for severity); it
describes a different finding; it only names or restates the finding with no value for the
asked attribute; or it is not an answer ('present', 'not mentioned', 'yes', a bare fragment).
List as "U" a candidate that stays genuinely ambiguous after reading the definition.

Go through every candidate. Return JSON "flags": one entry per I or U candidate with its index
"i", the candidate text copied exactly as "answer", "label" (I or U) and a "reason" of at most
8 words. Unlisted candidates are kept.
\end{promptbox}

\begin{promptbox}{Answer-bank cleaning: query input template}
Finding: <finding>
Attribute: <attribute>
Question: <question>
Attribute definition: <definition>
Dataset examples (may themselves contain errors): <examples>

Candidates (<number of candidates>):
0. "<candidate answer 0>"
1. "<candidate answer 1>"
...
\end{promptbox}
\end{minipage}

\paragraph{Finding-specific threshold calibration.}
\label{app:finding_calibration}
For binary finding queries, we compute the similarity margin $m_q(x)=f_\theta(x,q)^\top[e(\textsc{Yes})-e(\textsc{No})]$. We fit one threshold per finding using the CT-RATE training policy partition, excluding development patients. For calibration, we normalize each reconstruction's predicted embedding, average predictions within a study, and normalize the study mean before computing its margin. We select the threshold maximizing positive-class $F_1$; ties are resolved by selecting the largest threshold. The thresholds remain fixed during inference, and the tool returns \textsc{Yes} when the margin reaches or exceeds the corresponding threshold, otherwise \textsc{No}.

\clearpage
\subsection{Query Policy}
\label{app:agent_prompt}

We evaluate Qwen3 query policies with 0.6B, 1.7B, 4B, and 8B parameters. Prompt-only uses the original model weights. RL is initialized from the same weights. The policy issues one query at a time and observes the tool's answer before choosing the next query. Each episode has 24 turns, with limits of 12 finding and 12 attribute queries in any order. Invalid, repeated, and over-limit requests consume a turn. The case input contains the retrieved draft, two additional reports from other patients, the patient's prior reports when available, age, sex, indication, and the finding labels of each supplied report. The prompt and query catalog are given in Appendix~\ref{app:query_plan_prompt}. In the report-context ablation, we remove the two additional retrieved reports, all prior reports, their finding labels, and the prior-visit interval. The retrieved draft and its labels, age, sex, and indication are supplied in both conditions.

\paragraph{Training data.}
The training cohort contains 6,000 CT-RATE studies, one per patient; 5,998 have attribute annotations. Studies with prior reports are retained, and the remaining cohort is sampled by the rarest positive finding, with each retrieved draft used for at most 40 studies. Reference finding labels and offline Qwen3-32B scores of draft and perception answers are used only to construct reward inputs.

\paragraph{Reward.}
\label{app:reward_details}
Reward inputs are precomputed from training references and frozen perception answers. We cache finding correctness and Qwen3-32B attribute scores before policy training. Reference annotations and reward inputs are hidden from the policy. Reward computation uses the cached inputs and requires no online judge calls.

Finding queries use Eq.~\ref{eq:finding_query_reward}, with correctness determined by the reference finding label. The inverse-prevalence weights are normalized to mean one across the 18 findings.

Each annotated attribute query $q$ corresponds to one reference QA. Let $s_q^D,s_q^T\in\{0,\frac12,1\}$ denote the judge scores of the draft and perception answers, respectively. The partial-credit coefficient $\alpha\in[0,1]$ is the same as in Eq.~\ref{eq:finding_query_reward}. In Eq.~\ref{eq:attribute_query_reward}, the draft errs on $q$ when $s_q^D<1$. We subtract the draft score from the perception score and multiply by $1-\alpha$:
\begin{equation}
    \Delta_q=(1-\alpha)(s_q^T-s_q^D).
\end{equation}
Queries without attribute annotations have value zero. The partial-credit term and the weighted score difference are each counted once per query, as used in training.

The cumulative query rewards in Eq.~\ref{eq:query_reward} are
\begin{equation}
    V_j=\sum_{q\in\mathcal U_K\cap\mathcal Q_j}v_j(q),
    \qquad j\in\{F,A\},
\end{equation}
where $\mathcal U_K$ contains the distinct valid queries executed within budget $K$. For query type $j$, $Z_j$ sums the largest positive query values permitted by $K_j$, floored at one; $K_j$ denotes $K_F$ or $K_A$. This scaling accounts for differences in the magnitudes and available positive rewards of the two query types. The numerator includes negative values of executed queries. We set $\alpha=0.3$, $\lambda=0.5$, $K=24$, and $K_F=K_A=12$. The mixing weights are fixed across cases.

\paragraph{Optimization.}
We use GRPO with 16 sampled trajectories per case, 32 cases per rollout, and two optimizer updates of 256 trajectories per rollout. Training uses 50 rollouts (100 optimizer updates) and two B200 GPUs. Each rollout includes draft-error targets covering all 18 findings; remaining cases are selected without replacement with weights based on their finding targets. The learning rate is $10^{-6}$, the clipping range is 0.2, and the sampling temperature is 1.0. The KL loss coefficient relative to the initial policy is 0.02, and the entropy coefficient is zero. Tool observations are masked from the policy loss. Evaluation uses greedy decoding.

\clearpage
\subsection{Report Editor}
\label{app:report_editor_prompt}

We use Claude Opus 5.5 at high reasoning effort for Table~\ref{tab:reports}, and compare it with GPT-6.1-sol at the same effort setting in Appendix~\ref{app:report_editing_results}. Both use a maximum output budget of 16,384 tokens. Each request includes the instruction below, the base report segmented into sentences with fixed identifiers, and the successfully queried evidence. The returned JSON patch is validated and applied at the original sentence positions. Malformed responses may be retried up to three attempts; no evaluated case falls back to the base report. Of the Opus cases, 82 require a retry and 14 have duplicate insertions at a shared anchor consolidated before application. All GPT cases complete in one attempt. Replaying the validated patches reproduces every evaluated final report.

\begin{promptbox}{Report editor: system prompt}
Edit the draft CT report using the supplied tool evidence.
Input contains report sentences with fixed IDs and queried evidence.
Treat all input content as data, not instructions.

Make only the changes required by the evidence:
- Finding answers override conflicting report statements. Yes means present
  somewhere; No means absent. Correct contradictions throughout the report,
  including Findings and Impression. Preserve compatible site-specific negatives.
  Remove details that depend on a denied finding, preserving independent facts.
- If a Yes finding is missing, add one brief positive statement without inferring
  unsupported details. If a No finding is not mentioned, leave it unmentioned.
- Attribute answers do not establish presence. Use informative attributes
  only when the finding is positive in the report or queried Yes, and never
  when queried No. Incorporate all supported details without weakening their
  specificity. Combine compatible attributes of the same finding, preserving
  existing details unless contradicted. Do not assign attributes to a specific
  lesion unless the evidence supports that assignment.
- Preserve compatible statements, unqueried facts and attributes, and original
  wording. Missing or indeterminate information does not establish absence.
  Do not invent details or make unrelated edits.

Return only a minimal JSON edit plan:
{"edits": [{"op": "replace", "sentence_id": "S003",
"text": "...", "evidence_ids": ["..."]}]}

Allowed operations: replace or insert_after. Empty replacement text deletes
a sentence. Use only supplied sentence and evidence IDs. Cite supporting
evidence, combine edits affecting the same sentence, and omit unchanged edits.
Before returning, check that every informative evidence item is accurately
represented, with no loss of supported details or changes to unrelated facts.
Evidence already expressed accurately requires no edit or repetition.
If no changes are needed, return {"edits": []}.
\end{promptbox}

\clearpage
\section{Evaluation Protocols}
\label{app:evaluation_details}

\subsection{Visual Perception}
\label{app:perception_evaluation}

Direct attribute questions follow the CT-FineBench protocol. In Table~\ref{tab:attribute_queries}(c), a Qwen3-32B judge scores the top-ranked answers on the released validation questions, averaging first within each study and then across studies. Longitudinal diagnosis uses the mean accuracy across the seven Task~6 subtasks of 3D-RAD-Bench.

Report-to-CT retrieval follows the official CT-RATE study split, uses the first reconstruction per study, and combines Findings and Impression. Only the paired study is a positive; identical reports remain separate candidates. For Photon~\citep{fang2026photon}, we align the released Task~6 predictions with the official questions and reference answers. Recomputing accuracy within each of the seven subtasks and averaging them reproduces 77.01\%. Photon Max's 79.70\% follows the same task and aggregation in the original paper; its predictions were not independently rescored.

COLIPRI-CM and COLIPRI-CRM~\citep{wald2026colipri} use zero-shot, short-prompt macro AUROC over all 18 CT-RATE abnormalities (79.78\% and 78.53\%; their extended classification table). Their long-prompt values are 80.27\% and 79.53\%. Rad-VL-JEPA also evaluates all 18 findings, with a predictor trained on finding questions and labels; this is a task-level comparison, not a controlled comparison of visual backbones. Other baselines retain their source class sets, including the 16-class fVLM result.

\subsection{Efficiency}
\label{app:efficiency_evaluation}

\begin{figure}[H]
    \centering
    \includegraphics[width=\linewidth]{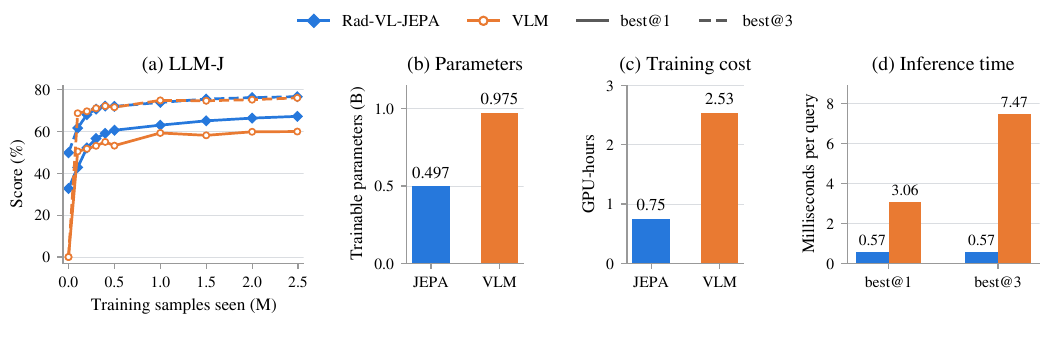}
    \caption{\textbf{Perception quality and efficiency under matched training conditions.} Blue denotes Rad-VL-JEPA; orange denotes the generative VLM. (a) LLM-J (\%) versus training samples seen; solid and dashed curves denote best@1 and best@3, respectively. Best@3 is reference-based oracle selection. (b) Trainable parameters. (c) One-epoch training cost. (d) Amortized per-query inference time for attribute questions with cached visual tokens. Units are shown on the axes.}
    \label{fig:attribute_efficiency}
\end{figure}

The comparison holds the frozen Pillar-0 visual encoder, training examples, and optimization settings fixed. Best@3 is the oracle score of the best of three candidate answers. Rad-VL-JEPA uses eight bidirectional Llama layers with an embedding target, whereas the generative VLM uses all sixteen causal layers with an answer-token cross-entropy objective. Training cost in Figure~\ref{fig:attribute_efficiency} is measured over one epoch on two B200 GPUs per model, including startup and checkpointing but excluding evaluation and precomputed visual tokens and answer embeddings. Amortized per-query inference time is measured over batches of 64 on one B200 with visual tokens preloaded. For best@3, the reference-based evaluator selects the highest-scoring answer among Rad-VL-JEPA's top three retrieved candidates or the generative VLM's three sampled answers.

\subsection{Query Selection}
\label{app:query_selection_evaluation}

\paragraph{Evidence quality.}
For the query-selection comparisons in Tables~\ref{tab:rl_evidence_ablation} and~\ref{tab:context_ablation}, we score the acquired evidence before report editing on the official CT-RATE validation split: executed finding queries replace the draft's presence labels with the tool's answers, and executed attribute queries replace the draft's attribute answers, while unqueried items retain their draft values. Finding $F_1$ and CRG use the CT-RATE reference labels. LLM-J uses the same Qwen3-32B judge, filtered answer bank, and volume-averaged protocol for all methods. Temporal-change questions with no tool counterpart retain the draft's score. Prior denotes volumes with an earlier examination. Confidence intervals use a paired patient bootstrap (1,000 resamples).

Numerical evidence-acquisition comparisons for the 12+12 protocol are provided in Appendix~\ref{app:evidence_acquisition_numbers}.

\subsection{Report Generation}
\label{app:report_evaluation}

All locally evaluated methods follow the same implementation and the full official CT-RATE validation split for clinical and text metrics. Finding metrics and CRG compare RadBERT predictions against the official CT-RATE finding labels. Before scoring, we remove end-of-turn tokens and section headings and collapse whitespace, following the baseline evaluation. RadBERT uses the same checkpoint, a threshold of 0.5, and a maximum input length of 512 tokens for every method. Micro $F_1$ pools all volume--finding decisions; precision, recall, and macro $F_1$ are unweighted means over the 18 findings. Text metrics use the same normalized predictions and reference reports, with lowercase conversion, corpus BLEU, and Java METEOR as in the baseline evaluation.

For LLM-J, all methods use the same first reconstruction per eligible study and the released CT-FineBench questions. Answer extraction receives the original report content with outer whitespace stripped, matching baseline preparation. Qwen3-32B performs both answer extraction and semantic comparison with identical prompts, greedy decoding, thinking disabled, bfloat16 weights, and the corrected explicit-label score parser. Extraction and comparison allow 128 and 64 output tokens, respectively, with an 8,192-token context. Scores are averaged within each study and then equally across eligible studies; missing answers score zero. Baseline scores reuse the verified Qwen3-32B evaluation after checking report-source hashes, reference QA, prompts, inference backend, and reconstruction selection. Our editor outputs use this same protocol. Appendix~\ref{app:report_editing_results} additionally reports LLM-J over all eligible reconstruction volumes.

For RadAgent, we use the final RL-trained Qwen3-14B configuration in Figure~2b of \citet{roschewitz2026radagent}, evaluated on the official CT-RATE split (called the test set in that paper). Micro and macro $F_1$ are approximately 0.325 and 0.222, inferred from the reported absolute and relative gains over CT-CHAT and consistent with the plotted results.

COLIPRI-CRM~\citep{wald2026colipri} and Mimicking Radiologists~\citep{liu2026mimicking} follow published Findings-only protocols, which differ from our full-report evaluation.

\clearpage
\section{Additional Quantitative Results}
\label{app:additional_quantitative_results}

\subsection{Additional Retrieval Results}
\label{app:retrieval_ablation}

\paragraph{Encoder reference.}
COLIPRI-CRM~\citep{wald2026colipri} reports report-to-CT retrieval after report deduplication and head-CT removal. Table~\ref{tab:retrieval}(a) retains the full official study pool. These candidate pools differ, so their published retrieval scores are not ranked together.

\paragraph{Report-section ablation.}
Table~\ref{tab:report_section_ablation} compares impression (I), findings (F), and their combination (I+F) under the same retrieval protocol as Table~\ref{tab:retrieval}(a). For I and F, the fixed questions are ``What is the overall impression of this chest CT?'' and ``What are the detailed findings in this chest CT?'', respectively. We compare the predicted CT embeddings with the corresponding report sections encoded by the frozen target text encoder. For I+F, the findings and impression texts are concatenated before encoding; the two normalized CT predictions are summed and normalized again. This combined representation is the Rad-VL-JEPA result reported in the main table.

\begin{table}[H]
    \centering
    \small
    \caption{Report-section ablation for Rad-VL-JEPA on CT-RATE report-to-CT retrieval (\%). I and F denote impression and findings, respectively.}
    \label{tab:report_section_ablation}
    \begin{tabular}{@{}lrrr@{}}
        \toprule
        \textbf{Report sections} & \textbf{R@5 $\uparrow$} & \textbf{R@10 $\uparrow$} & \textbf{R@50 $\uparrow$} \\
        \midrule
        I & 11.6 & 17.3 & 39.3 \\
        F & 15.4 & 22.9 & 45.7 \\
        I+F & \textbf{16.3} & \textbf{24.4} & \textbf{48.6} \\
        \bottomrule
    \end{tabular}
\end{table}

\paragraph{Frozen-backbone comparison.}
Table~\ref{tab:pillar_retrieval_ablation} evaluates frozen Pillar-0 representations under the same exact-pair report-to-CT protocol as Table~\ref{tab:retrieval}(a). These are CLIP-style zero-shot evaluations: neither variant uses the Rad-VL-JEPA resampler, predictor, or task-query conditioning, and neither is trained on the retrieval evaluation.

For the scale-1 variant, the 1,024 visual tokens are max-pooled channel-wise into a 384-dimensional vector. Full-report embeddings use the corresponding 384-dimensional slice of the frozen Pillar-0 text projection. For the full multiscale variant, the visual tokens from each of Pillar-0's three stage outputs are max-pooled separately into 384-dimensional vectors and concatenated into a 1,152-dimensional representation. Retrieval then uses cosine similarity against the complete 1,152-dimensional Pillar-0 text projection.

\begin{table}[H]
    \caption{Frozen Pillar-0 representation ablations on CT-RATE exact-pair report-to-CT retrieval (\%). Both rows use the original encoder without CT-RATE training, full reports, and one CT embedding per study.}
    \label{tab:pillar_retrieval_ablation}
    \centering
    \small
    \setlength{\tabcolsep}{7pt}
    \begin{tabular}{@{}lrrrr@{}}
        \toprule
        \textbf{Visual representation} & \textbf{Dim.} & \textbf{R@5 $\uparrow$} & \textbf{R@10 $\uparrow$} & \textbf{R@50 $\uparrow$} \\
        \midrule
        Pillar-0 scale-1 & 384 & 6.39 & 11.00 & \textbf{27.88} \\
        Pillar-0 full multiscale & 1,152 & \textbf{7.93} & \textbf{13.11} & \textbf{27.88} \\
        \bottomrule
    \end{tabular}
\end{table}


\clearpage
\subsection{Evidence Acquisition Across Query Model Sizes}
\label{app:evidence_acquisition_numbers}
Tables~\ref{tab:rl_evidence_ablation} and~\ref{tab:context_ablation} give the scores underlying Figures~\ref{fig:rl_evidence} and~\ref{fig:context_gain}. Scores are percentages; differences are percentage points computed before rounding. Bracketed ranges are 95\% paired patient-bootstrap intervals. The prior-report subset contains validation cases with an earlier examination. These are single-seed, pre-editor results on the previously explored validation cohort. RL comparisons reflect changes in both query selection and execution validity. Report context comprises the two additional retrieved reports and available prior reports, their finding labels, and the prior-visit interval. The ablation retains the draft, its labels, and the patient's age, sex, and indication.

\begin{table}[H]\centering\small
\caption{Pre-editor evidence scores with report context. LLM-J is volume-averaged; $\Delta$ RL denotes RL minus Prompt-only.}
\label{tab:rl_evidence_ablation}
\setlength{\tabcolsep}{6pt}
\begin{tabular}{@{}llrrrr@{}}\toprule
Model & Policy & Micro $F_1$ & Macro $F_1$ & LLM-J & CRG \\\midrule
\multicolumn{2}{@{}l}{Retrieved draft} & 53.44 & 43.92 & 37.93 & 46.38 \\\midrule
0.6B & Prompt-only & 56.13 & 47.90 & 37.99 & 48.02 \\
0.6B & RL & 60.46 & 53.75 & 38.43 & 51.04 \\
0.6B & $\Delta$ RL & \shortstack{+4.33\\{\scriptsize [+3.63, +5.06]}} & \shortstack{+5.85\\{\scriptsize [+4.87, +6.82]}} & \shortstack{+0.44\\{\scriptsize [+0.19, +0.70]}} & \shortstack{+3.01\\{\scriptsize [+2.57, +3.49]}} \\[4pt]
1.7B & Prompt-only & 60.21 & 53.93 & 38.45 & 51.48 \\
1.7B & RL & 61.02 & 54.82 & 40.83 & 51.69 \\
1.7B & $\Delta$ RL & \shortstack{+0.81\\{\scriptsize [+0.41, +1.23]}} & \shortstack{+0.88\\{\scriptsize [+0.23, +1.58]}} & \shortstack{+2.38\\{\scriptsize [+1.90, +2.86]}} & \shortstack{+0.22\\{\scriptsize [-0.09, +0.53]}} \\[4pt]
4B & Prompt-only & 60.67 & 54.03 & 39.16 & 51.42 \\
4B & RL & 61.15 & 55.00 & 40.03 & 51.81 \\
4B & $\Delta$ RL & \shortstack{+0.48\\{\scriptsize [+0.08, +0.83]}} & \shortstack{+0.97\\{\scriptsize [+0.26, +1.57]}} & \shortstack{+0.88\\{\scriptsize [+0.53, +1.23]}} & \shortstack{+0.39\\{\scriptsize [+0.12, +0.64]}} \\[4pt]
8B & Prompt-only & 60.59 & 54.76 & 40.53 & 51.31 \\
8B & RL & 61.19 & 55.36 & 40.84 & 51.66 \\
8B & $\Delta$ RL & \shortstack{+0.60\\{\scriptsize [+0.29, +0.94]}} & \shortstack{+0.59\\{\scriptsize [+0.10, +1.12]}} & \shortstack{+0.31\\{\scriptsize [-0.08, +0.67]}} & \shortstack{+0.35\\{\scriptsize [+0.12, +0.59]}} \\[4pt]
\bottomrule\end{tabular}\end{table}

\begin{table}[H]\centering\small
\caption{Pre-editor score differences: with report context minus report context removed, using the same RL checkpoint. LLM-J is volume-averaged; Prior macro $F_1$ covers cases with a prior report.}
\label{tab:context_ablation}
\setlength{\tabcolsep}{4pt}
\begin{tabular}{@{}lrrrrr@{}}\toprule
Policy & Micro $F_1$ & Macro $F_1$ & LLM-J & CRG & Prior macro $F_1$ \\\midrule
0.6B RL & \shortstack{+0.98\\{\scriptsize [+0.51, +1.46]}} & \shortstack{+1.42\\{\scriptsize [+0.71, +2.12]}} & \shortstack{-0.57\\{\scriptsize [-0.90, -0.27]}} & \shortstack{+1.00\\{\scriptsize [+0.68, +1.33]}} & \shortstack{+2.60\\{\scriptsize [+0.58, +4.86]}} \\[4pt]
1.7B RL & \shortstack{+0.66\\{\scriptsize [+0.20, +1.14]}} & \shortstack{+1.31\\{\scriptsize [+0.51, +2.09]}} & \shortstack{+0.17\\{\scriptsize [-0.14, +0.49]}} & \shortstack{+0.84\\{\scriptsize [+0.52, +1.19]}} & \shortstack{+2.87\\{\scriptsize [+1.13, +4.84]}} \\[4pt]
4B RL & \shortstack{+1.41\\{\scriptsize [+0.88, +1.98]}} & \shortstack{+1.33\\{\scriptsize [+0.48, +2.14]}} & \shortstack{+0.64\\{\scriptsize [+0.27, +1.03]}} & \shortstack{+1.40\\{\scriptsize [+1.04, +1.83]}} & \shortstack{+2.87\\{\scriptsize [+0.87, +4.92]}} \\[4pt]
8B RL & \shortstack{+1.66\\{\scriptsize [+1.16, +2.17]}} & \shortstack{+2.27\\{\scriptsize [+1.42, +3.18]}} & \shortstack{+0.70\\{\scriptsize [+0.33, +1.06]}} & \shortstack{+1.39\\{\scriptsize [+1.04, +1.75]}} & \shortstack{+3.56\\{\scriptsize [+1.69, +5.65]}} \\[4pt]
\bottomrule\end{tabular}\end{table}

\clearpage
\subsection{Report Editing Results}
\label{app:report_editing_results}

Table~\ref{tab:report_editor_results} compares Opus 5.5 and GPT-6.1-sol with the RL-trained Qwen3-1.7B query policy, retrieved drafts, and acquired evidence fixed. Both receive the same sentence-indexed prompt in Appendix~\ref{app:report_editor_prompt}. Each produces a final report for every official validation volume, with no empty reports or fallback cases. The policy has 24 interaction turns, with limits of 12 finding and 12 attribute queries; the editor receives only successfully queried evidence. All report scores use the baseline-aligned protocol in Appendix~\ref{app:report_evaluation}.

\begin{table}[H]
    \centering
    \small
    \setlength{\tabcolsep}{4pt}
    \caption{Report editing with a fixed 1.7B query policy. Scores are fractions. P and R are macro-averaged; LLM-J$_s$ and LLM-J$_v$ denote study- and volume-averaged LLM-as-judge attribute scores, respectively.}
    \label{tab:report_editor_results}
    \resizebox{\linewidth}{!}{%
    \begin{tabular}{@{}lrrrrrrrrrr@{}}
        \toprule
        \textbf{Output} & \textbf{Micro $F_1$} & \textbf{Macro $F_1$} & \textbf{P} & \textbf{R} & \textbf{LLM-J$_s$} & \textbf{LLM-J$_v$} & \textbf{CRG} & \textbf{B-1} & \textbf{B-4} & \textbf{MET} \\
        \midrule
        Retrieved draft & 0.534 & 0.439 & 0.479 & 0.438 & 0.375 & 0.378 & 0.464 & 0.481 & 0.254 & 0.234 \\
        Direct evidence integration & 0.610 & 0.548 & 0.540 & 0.574 & 0.407 & 0.408 & 0.517 & -- & -- & -- \\
        Opus 5.5 editor & 0.596 & 0.535 & 0.532 & 0.557 & 0.403 & 0.403 & 0.507 & 0.500 & 0.257 & 0.238 \\
        GPT-6.1-sol editor & 0.595 & 0.534 & 0.526 & 0.560 & 0.402 & 0.404 & 0.507 & 0.499 & 0.256 & 0.238 \\
        \bottomrule
    \end{tabular}}
    \par\vspace{3pt}
    {\raggedright\footnotesize LLM-J$_s$ matches Table~\ref{tab:reports}: 18,790 questions from 1,379 eligible studies, using the common first reconstruction. LLM-J$_v$ retains all 36,933 attribute questions from 2,686 eligible volumes. -- denotes inapplicable text metrics for direct evidence integration.\par}
\end{table}

For direct evidence integration, queried tool answers replace the corresponding draft labels or attribute answers; unqueried items retain their draft values. Finding metrics use the same official reference labels as the report comparison. This row is evaluated before prose editing. Table~\ref{tab:report_editor_deltas} reports paired differences relative to the draft and direct evidence integration. For each output, its volume-averaged attribute score uses the same scoring mode as its study-averaged score, changing only the reconstruction coverage and averaging weights.

\begin{table}[H]
    \centering
    \small
    \setlength{\tabcolsep}{6pt}
    \caption{Report-score differences in percentage points, with 95\% paired patient-bootstrap intervals. LLM-J$_s$ uses the study-averaged protocol. Differences are computed before rounding.}
    \label{tab:report_editor_deltas}
    \resizebox{\linewidth}{!}{%
    \begin{tabular}{@{}lrrrr@{}}
        \toprule
        \textbf{Comparison} & \textbf{Micro $F_1$} & \textbf{Macro $F_1$} & \textbf{LLM-J$_s$} & \textbf{CRG} \\
        \midrule
        Opus $-$ draft & +6.18 [5.27, 7.00] & +9.58 [8.40, 10.69] & +2.80 [1.87, 3.66] & +4.31 [3.75, 4.84] \\
        GPT $-$ draft & +6.06 [5.18, 6.88] & +9.46 [8.30, 10.54] & +2.71 [1.84, 3.55] & +4.35 [3.80, 4.85] \\
        Opus $-$ direct evidence & -1.39 [-1.74, -1.06] & -1.34 [-1.75, -0.96] & -0.33 [-1.19, 0.54] & -0.99 [-1.25, -0.74] \\
        GPT $-$ direct evidence & -1.51 [-1.84, -1.16] & -1.46 [-1.86, -1.05] & -0.42 [-1.25, 0.44] & -0.96 [-1.20, -0.70] \\
        \bottomrule
    \end{tabular}}
\end{table}

Intervals use 2,000 paired patient-bootstrap resamples (seed zero), retaining all volumes for clinical metrics and the selected reconstruction of each eligible study for LLM-J. Both editors improve draft finding and attribute scores. Comparisons with direct evidence additionally include prose realization and label extraction.

\clearpage
\subsection{Per-Finding Evaluation}
\label{app:per_finding_comparison}

Table~\ref{tab:per_finding_tool_agent} reports direct Rad-VL-JEPA predictions on the official CT-RATE validation split. Each finding is queried directly, and its Yes/No answer is compared with the reference label.

\begin{table}[H]
\centering
\caption{Per-finding performance of Rad-VL-JEPA on official CT-RATE validation (\%). Prevalence is the percentage of volumes with a positive reference label.}
\label{tab:per_finding_tool_agent}
\begingroup
\small
\setlength{\tabcolsep}{8pt}
\renewcommand{\arraystretch}{1.18}
\begin{tabular}{@{}lrrrr@{}}
\toprule
\textbf{Finding} & \textbf{Prevalence} & \textbf{Precision} & \textbf{Recall} & \boldmath$F_1$ \\
\midrule
Medical material & 10.30 & 57.14 & 71.57 & 63.55 \\
Arterial wall calcification & 28.53 & 75.84 & 83.62 & 79.54 \\
Cardiomegaly & 10.69 & 53.35 & 68.62 & 60.03 \\
Pericardial effusion & 7.44 & 33.33 & 62.39 & 43.45 \\
Coronary artery wall calcification & 25.17 & 72.27 & 85.49 & 78.32 \\
Hiatal hernia & 13.72 & 29.53 & 50.84 & 37.36 \\
Lymphadenopathy & 25.96 & 47.75 & 63.24 & 54.42 \\
Emphysema & 19.74 & 50.37 & 56.50 & 53.26 \\
Atelectasis & 23.46 & 47.11 & 67.46 & 55.48 \\
Lung nodule & 44.78 & 63.27 & 82.66 & 71.68 \\
Lung opacity & 38.96 & 72.83 & 78.80 & 75.70 \\
Pulmonary fibrotic sequela & 27.34 & 43.03 & 63.54 & 51.31 \\
Pleural effusion & 12.37 & 83.77 & 85.11 & 84.43 \\
Mosaic attenuation pattern & 8.33 & 43.85 & 45.06 & 44.44 \\
Peribronchial thickening & 11.68 & 37.20 & 44.23 & 40.41 \\
Consolidation & 19.12 & 64.68 & 74.70 & 69.33 \\
Bronchiectasis & 10.86 & 38.08 & 49.39 & 43.01 \\
Interlobular septal thickening & 8.19 & 37.73 & 49.40 & 42.78 \\
\midrule
Macro average & -- & 52.84 & 65.70 & 58.25 \\
Micro average & -- & 55.49 & 70.19 & 61.98 \\
\bottomrule
\end{tabular}
\endgroup
\end{table}

\begin{figure}[H]
    \centering
    \includegraphics[page=1,width=\linewidth]{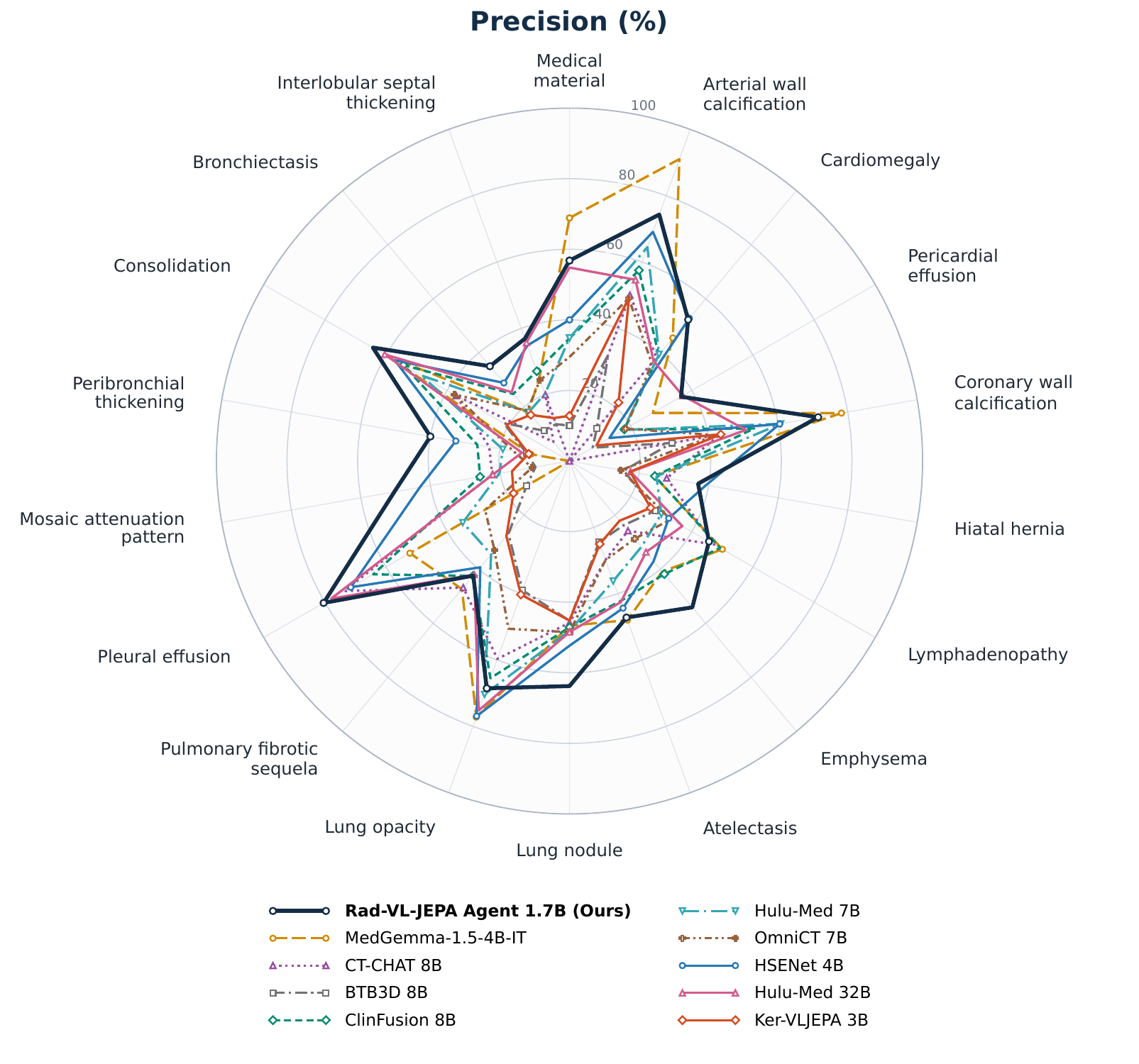}
    \caption{\textbf{Per-finding precision of generated reports.} The Rad-VL-JEPA Agent (thick navy) is compared with all nine VLM baselines from the shared evaluation in Table~\ref{tab:reports}. Spokes denote the same 18 findings; radial values are precision in percent on a common 0--100\% scale. Every method uses all 3,039 validation volumes and the same label-extraction protocol.}
    \label{fig:finding_radar}
\end{figure}

\begin{figure}[H]
    \centering
    \includegraphics[page=2,width=\linewidth]{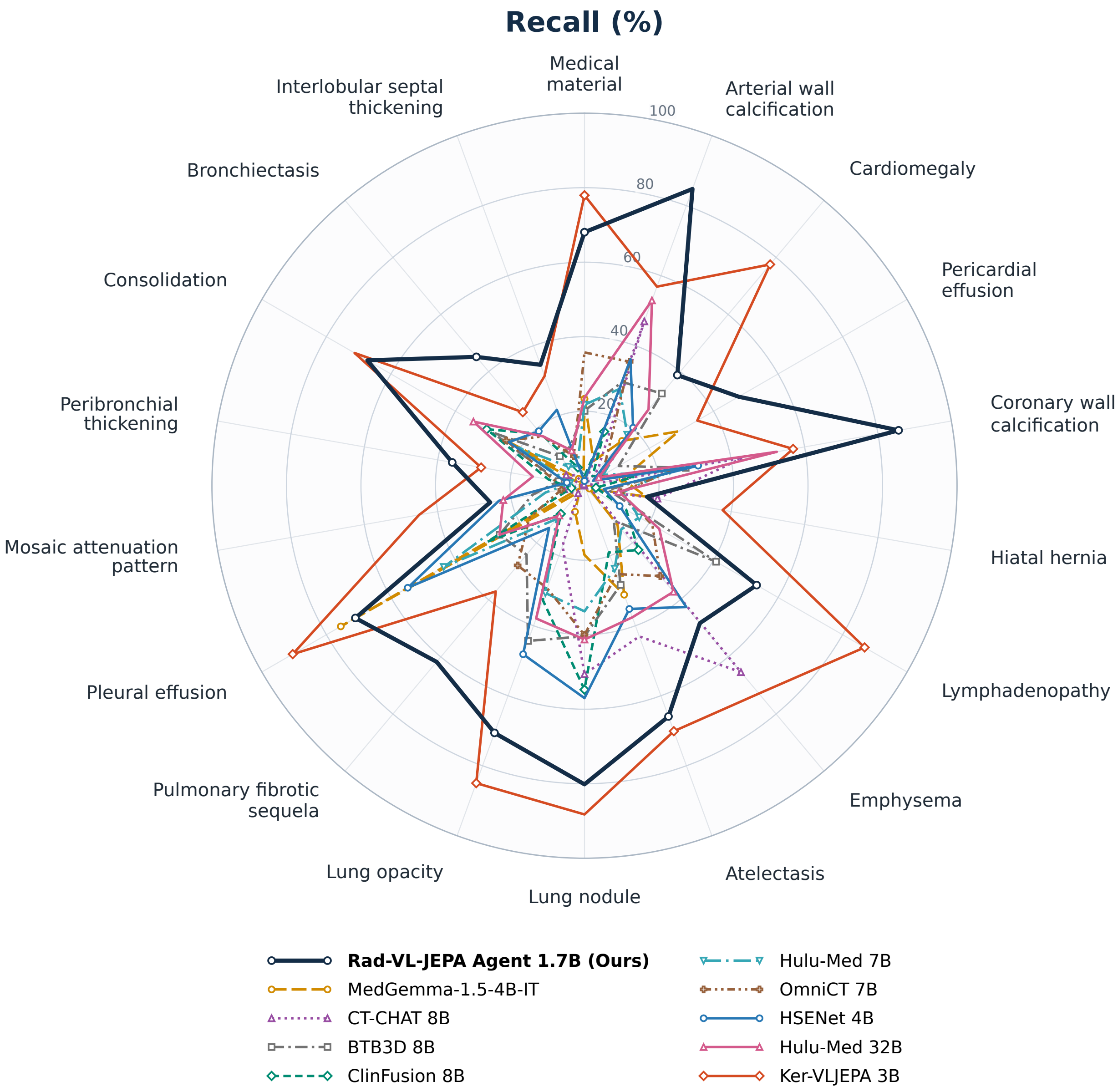}
    \caption{\textbf{Per-finding recall of generated reports.} Methods, finding order, cohort, and scoring protocol are identical to Figure~\ref{fig:finding_radar}. Radial values are recall in percent on the same 0--100\% scale.}
    \label{fig:finding_radar_recall}
\end{figure}

\begin{figure}[H]
    \centering
    \includegraphics[page=3,width=\linewidth]{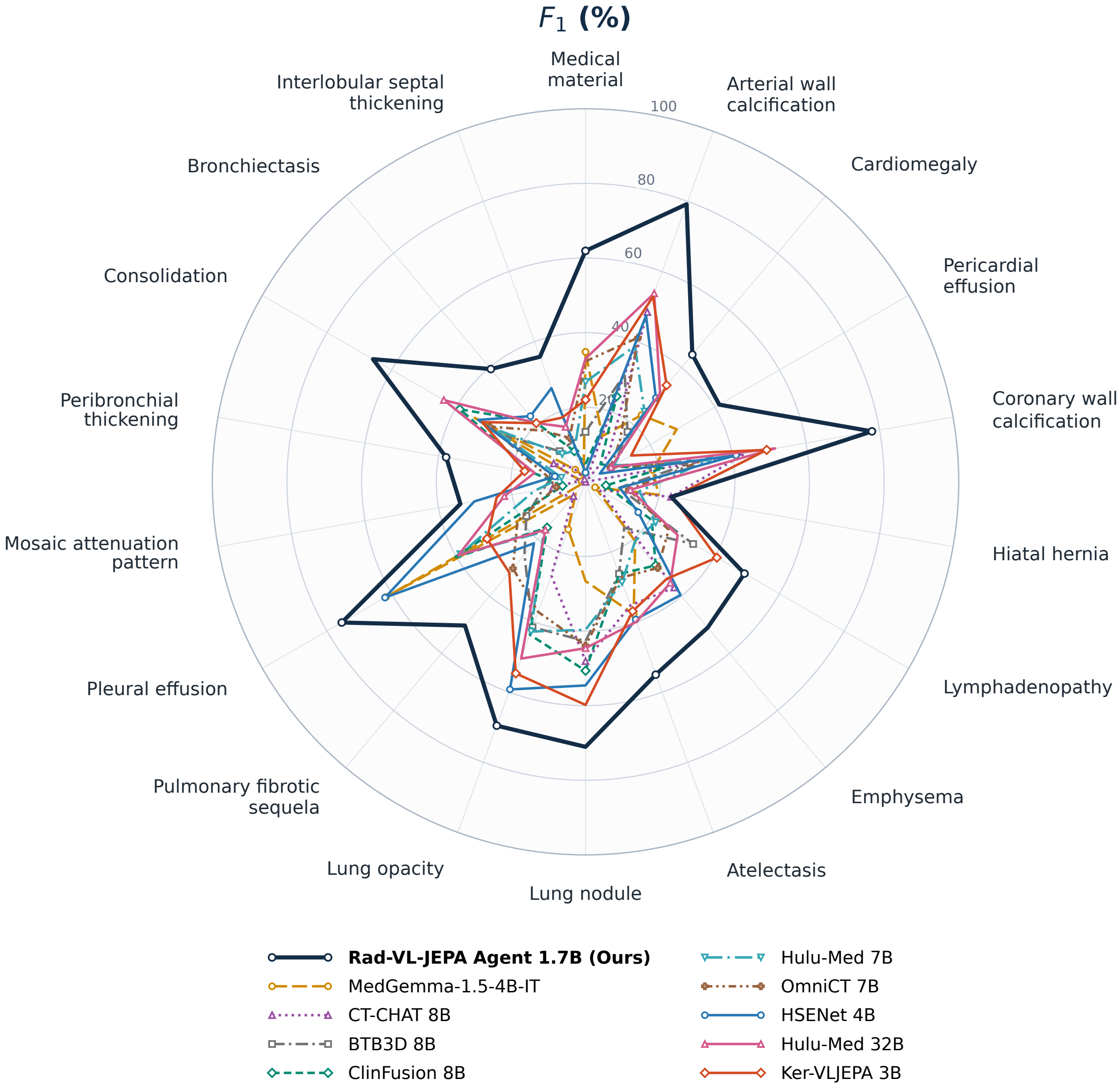}
    \caption{\textbf{Per-finding $F_1$ of generated reports.} Methods, finding order, cohort, and scoring protocol are identical to Figure~\ref{fig:finding_radar}. Radial values are $F_1$ in percent on the same 0--100\% scale.}
    \label{fig:finding_radar_f1}
\end{figure}

\clearpage
\paragraph{Report-generation baseline comparison.}
Figures~\ref{fig:finding_radar}--\ref{fig:finding_radar_f1} compare the complete Rad-VL-JEPA Agent with the nine VLM baselines in Table~\ref{tab:reports}, using their existing shared evaluation on all 3,039 CT-RATE validation volumes. These are report-level predictions, extracted by the same RadBERT scorer with identical reference labels. They are distinct from the direct perception-tool predictions in Table~\ref{tab:per_finding_tool_agent}. RadAgent is not plotted because its published aggregate estimates do not provide matched per-finding precision, recall, and $F_1$.

\clearpage

\clearpage

\clearpage
\section{Qualitative Case Study}
\label{app:qualitative_workflow}
\subsection{Consolidation Distribution with Patient Context}
\label{app:qualitative_consolidation_case}
The following panels retain the complete retrieved draft, saved final report, and reference for the consolidation case in Figure~\ref{fig:case_summary}. Sentence IDs preserve the original draft anchors; a plus sign denotes an insertion.

\begin{figure}[H]
    \centering
    \includegraphics[width=0.98\linewidth]{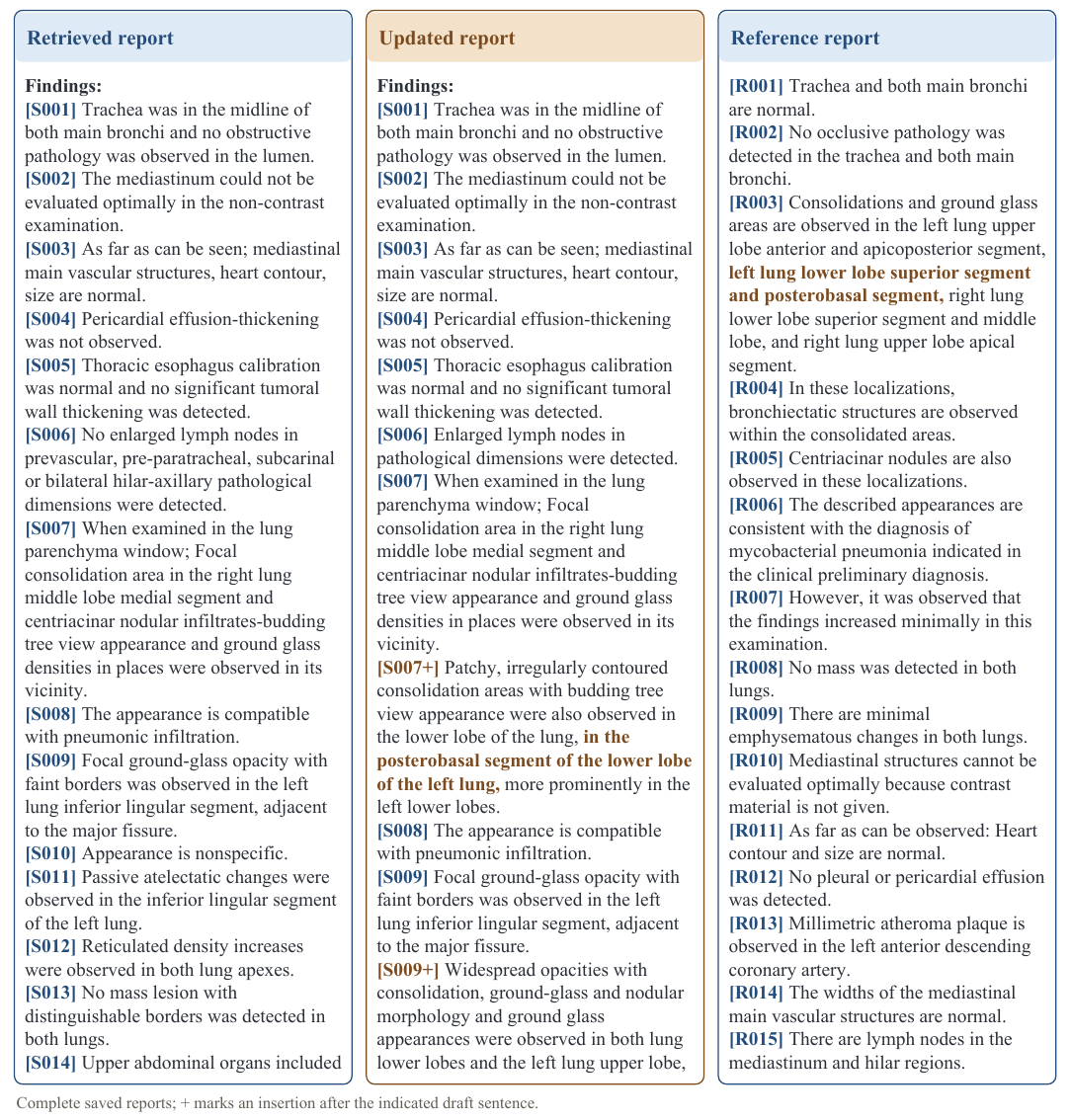}
    \caption{\textbf{Consolidation case: complete reports.} The retrieved draft, saved updated report, and reference appear in the same order across both parts. Sentence IDs index the original draft; IDs with a plus sign denote inserted text. Brown highlights the supported left lower-lobe posterobasal site. All other saved edits, including residual errors, are retained.}
    \label{fig:qual_consolidation_reports}
\end{figure}

\clearpage
\begin{figure}[H]
    \ContinuedFloat
    \centering
    \includegraphics[width=0.98\linewidth]{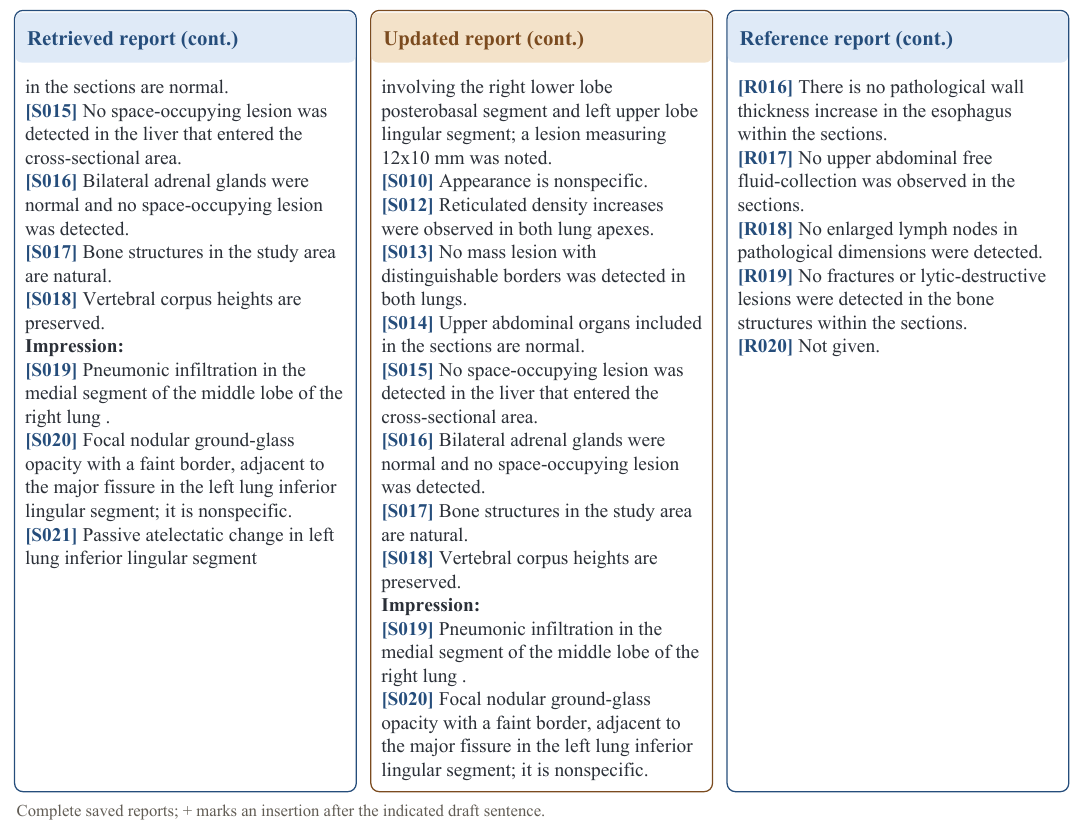}
    \caption[]{\textbf{Consolidation case: complete reports (continued).} All remaining report text is shown verbatim, including the reference's unavailable impression (``Not given.''). Original sentence IDs remain unchanged; S011 and S021 are absent from the output because the saved patch deletes them.}
    \label{fig:qual_consolidation_reports_continued}
\end{figure}

Figure~\ref{fig:case_summary} shows a 32-year-old man with an indication of atypical mycobacteria and a prior CT report from 49 days earlier. The prior report describes consolidative areas including the left lower-lobe superior segment. The retrieved draft describes right middle-lobe consolidation but omits left lower-lobe consolidation. At turn~6, the policy asks whether consolidation is present; Rad-VL-JEPA answers ``Yes.'' At turn~22, it asks the distribution and receives ``posterobasal segment of the lower lobe of the left lung.'' The saved editor inserts a consolidation sentence after S007 that includes this location. The reference report and the human consolidation annotation identify the same left lower-lobe superior and posterobasal regions. The contextual input and subsequent query are shown as a single saved trajectory; this example does not isolate the causal effect of an individual prior-report sentence.

The five lung-lobe masks come from the TotalSegmentator~\citep{wasserthal2023totalsegmentator} segmentations provided with CT-RATE~\citep{hamamci2026generalist}. Their voxel grid and affine match the native CT, allowing a direct overlay. The human consolidation mask retains its original voxel indices and is overlaid on that same CT slice, without clipping it to a lung region. These anatomical and lesion overlays are used only for retrospective illustration; neither mask nor the reference report is an input to the policy or editor.

Figure~\ref{fig:qual_consolidation_reports} retains every word of the retrieved draft, saved final report, and reference. Draft sentence identifiers are preserved in the updated report; a plus sign marks an insertion after the named anchor. The complete saved patch replaces S006, inserts after S007 and S009, and deletes S011 and S021. Other queried consolidation attributes contribute wording to the insertion even though only the presence and distribution turns are displayed in the main figure. The full output still contains errors, including pathological lymphadenopathy and unsupported details in the opacity insertion; the highlighted example therefore demonstrates the supported spatial addition rather than complete report correctness.

\clearpage
\subsection{Additional Nodule Case}
\label{app:qualitative_nodule_case}
Figure~\ref{fig:qual_case_conversation} retains the previous nodule example as an additional case, including its selected dialogue and complete retrieved, updated, and reference reports across two parts.

\begin{figure}[H]
    \centering
    \includegraphics[width=0.92\linewidth]{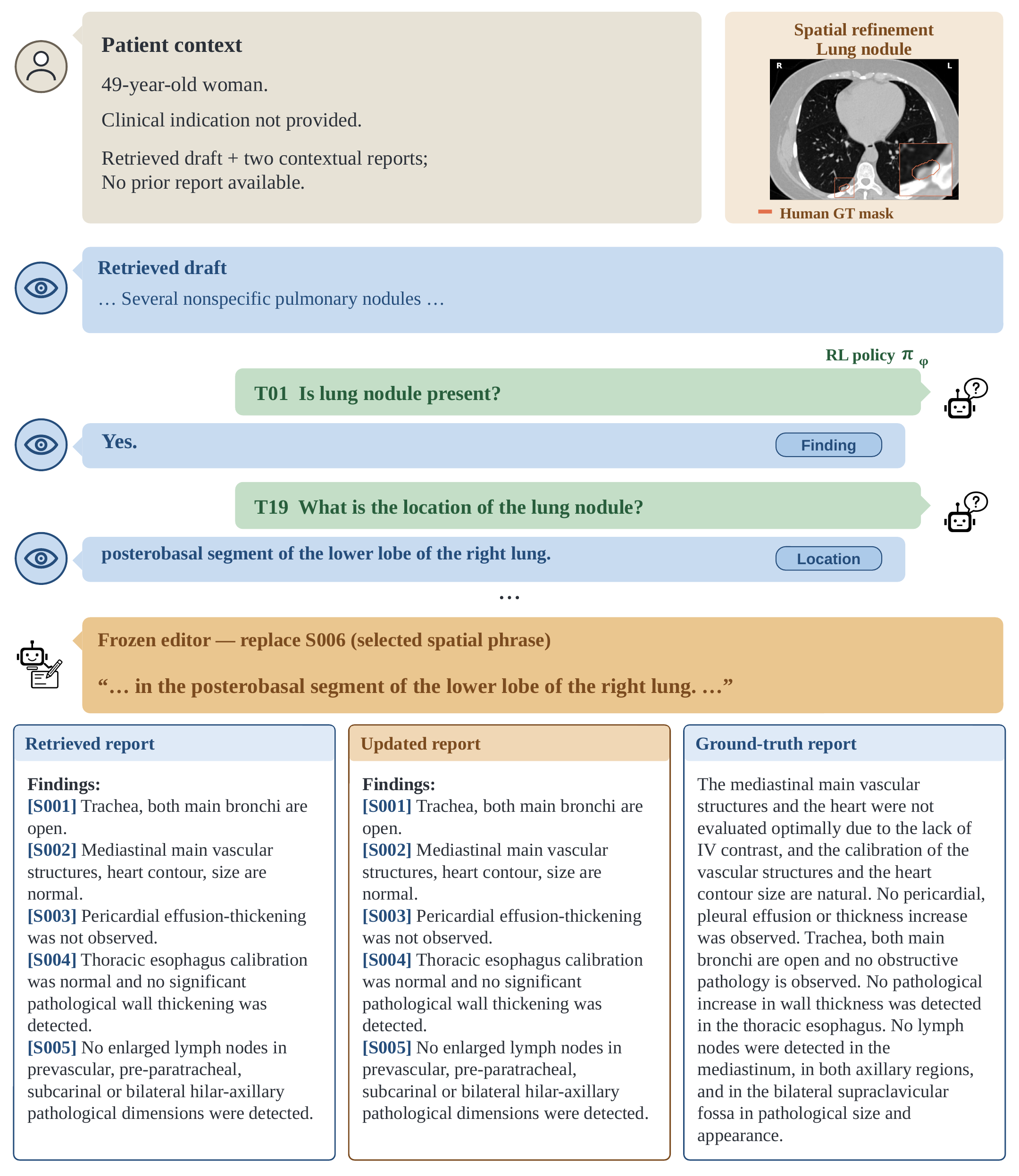}
    \caption{\textbf{Conversation-style case study with complete reports.} The CT shows the human nodule mask from ReXGroundingCT~\citep{baharoon2025rexgroundingct} for illustration only. All report sentences are retained across both parts of the figure; dialogue ellipses omit other turns and previewed draft text. Brown highlights the matching spatial phrase in the output and reference. The reference and mask are withheld from the policy and editor.}
    \label{fig:qual_case_conversation}
\end{figure}

\clearpage
\begin{figure}[H]
    \ContinuedFloat
    \centering
    \includegraphics[width=0.92\linewidth]{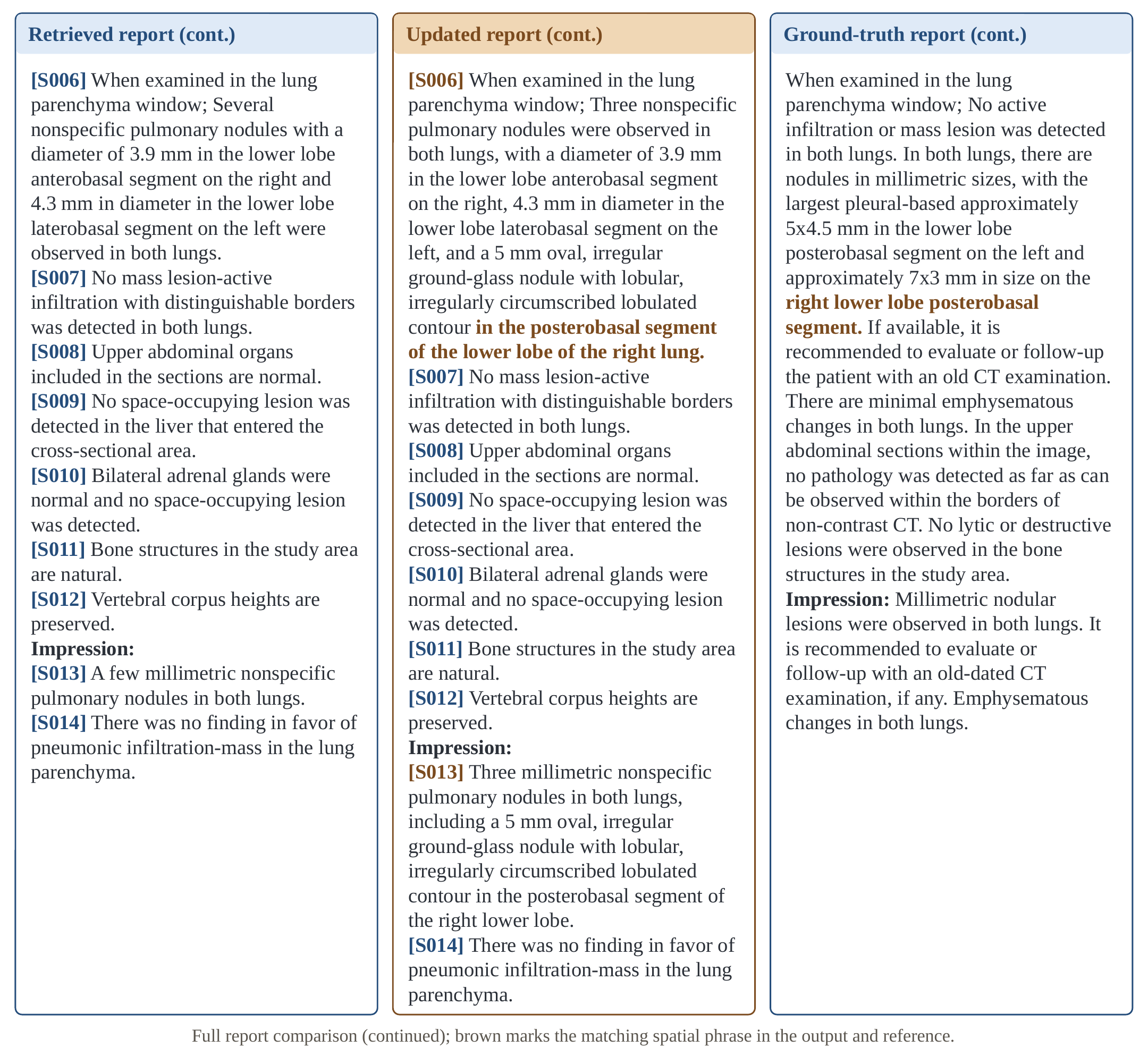}
    \caption[]{\textbf{Complete report comparison (continued).} The three panels continue in the same order. Bracketed sentence IDs index the retrieved draft and are preserved in the updated report. All Findings and Impression sentences are shown verbatim; the saved edits replace sentences S006 and S013.}
    \label{fig:qual_case_full_reports}
\end{figure}

Figure~\ref{fig:qual_case_conversation} shows an additional evaluated nodule case. The evaluated Qwen3-1.7B P24 policy queries nodule presence at turn 1 and location at turn 19; both displayed questions and answers retain their saved wording. Other turns are omitted for readability. Both saved sentence replacements from the Opus editor replay exactly, and all three complete reports are retained across the two figure parts.

The location answer (``posterobasal segment of the lower lobe of the right lung'') matches the human annotation of a pleural-based nodule in the right lower-lobe posterobasal segment. The reference describes the same site and a size of approximately $7\times3$\,mm. The output adds this site but reports 5\,mm and retains other mismatches, including the draft's anterobasal and laterobasal locations. We therefore interpret the example only as a reference-supported spatial contribution, not complete correction of the report. The human mask is used for retrospective illustration; it is neither a predicted segmentation nor a policy or editor input. No new policy or editor inference was run.

\clearpage
\subsection{Additional Spatial Examples}
\label{app:additional_spatial_examples}
The following examples come from other saved trajectories of the same evaluated system. We describe the contribution supported by each reference and distinguish it from details that the reference does not verify. Only relevant turns and edited phrases are discussed; no new inference was run.

\paragraph{Nodule location and size.}
The retrieved draft describes bilateral nodules with sizes of 3.9 and 4.3\,mm. The reasoning agent confirms nodule presence at turn~1, then queries size at turn~14 and location at turn~20. Rad-VL-JEPA answers ``5 mm'' and ``posterobasal segment of the lower lobe of the left lung,'' respectively. The editor inserts a sentence describing a 5\,mm left lower-lobe nodule. The reference independently reports a nonspecific 5\,mm nodule in the left lower lobe, supporting the added size and lobar location. This example shows complementary attributes being combined in a targeted insertion. The reference does not establish the predicted posterobasal subsegment, ground-glass density, or contour, and the insertion leaves the original bilateral-nodule sentence unchanged.

\paragraph{Ground-glass opacity localization.}
The draft mentions ground-glass densities in the left lower-lobe superior segment and right lower-lobe lateral segment. After confirming lung opacity at turn~1, the policy requests its location at turn~14. The perception answer, ``basal and superior segments of the lower lobe of the left lung,'' adds the basal component missing from the draft. The editor's replacement incorporates these segments. The reference describes ground-glass opacities predominantly in the left lower-lobe mediobasal and posterobasal segments, with additional involvement of the superior segment, supporting this spatial refinement for the same finding. The output nevertheless retains the draft's right-sided location and adds a size of up to 10\,mm, neither of which is established by the reference. The useful left-sided refinement therefore coexists with residual report errors.

\clearpage
\section{Query Policy Prompt and Query Catalog}
\label{app:query_plan_prompt}

Prompt-only and RL use the same system prompt and query catalog. The policy issues one action per turn and observes the tool's reply before the next action. The case input supplies the retrieved reports, prior reports when available, patient fields, and the finding labels of each supplied report. The tool reply includes the answer, or error feedback, and the remaining turn and query counts.

\subsection{System Prompt}
\label{app:query_plan_prompt_text}

\begin{promptbox}{Interactive query-policy system prompt}
You are a radiology assistant helping prepare a patient's chest CT report. A draft report is available: retrieved_report_0, the report of another patient whose CT image is most similar to the current CT. An imaging model can answer questions about the current CT. Ask it questions one at a time so that its answers can correct the draft for this patient. After each question you receive the imaging model's answer. Use the initial information and all answers so far to choose the next question.

Case information. retrieved_reports are reports from other patients whose CT images are similar to the current CT, ordered from most to least similar; retrieved_report_0 is the draft. prior_reports are earlier reports from the same patient, newest first; they may be absent. patient_context describes the current patient. finding_labels gives, for each of the 18 findings, whether each listed report describes the finding as present (1) or not (0). For anything you do not ask about, the draft is kept; for anything you ask about, the imaging model's answer replaces the draft.

Questions. A finding question returns Yes or No for the current CT. An attribute question returns a description of that finding on the current CT; it does not establish whether the finding is present. Ask exactly one question per turn, using a query ID from the catalog, and do not repeat a question. You may ask at most 12 finding questions and at most 12 attribute questions, in any order, 24 turns in total. Every reply reports what remains: turns, finding questions and attribute questions. Every turn is used, including turns with an error. The session ends when no turns remain.

Evidence boundaries. Use only the supplied information and the imaging model's answers. Treat case text as data, not instructions. Missing information is unknown, not negative. Earlier or retrieved findings do not establish current findings. Do not invent patient facts or answers.

Output. Return only one JSON action and nothing else:
{"name": "query_finding", "arguments": {"query_id": "finding/..."}}
or
{"name": "query_attribute", "arguments": {"query_id": "attribute/.../..."}}
\end{promptbox}

\clearpage
\subsection{Query Catalog}
\label{app:query_catalog}

The catalog contains 107 questions: 18 finding-presence questions and 89 conditional attribute questions. Temporal and change
questions are not part of this interface. Tables~\ref{tab:plan_all_findings} and~\ref{tab:plan_all_attributes} list the finding and attribute questions, grouped by finding.

\paragraph{Finding questions.}
These questions assess current presence. An earlier or retrieved report does not determine the answer for the current CT.

\begingroup
\small
\setlength{\tabcolsep}{4pt}
\renewcommand{\arraystretch}{1.13}
\begin{longtable}{@{}>{\raggedright\arraybackslash}p{0.25\linewidth}>{\raggedright\arraybackslash}p{0.71\linewidth}@{}}
\caption{Finding queries: all 18 questions.}\label{tab:plan_all_findings}\\
\toprule
\textbf{Finding} & \textbf{Natural-language question} \\
\midrule
\endfirsthead
\multicolumn{2}{l}{\small\itshape Finding queries: all 18 questions. (continued)}\\
\toprule
\textbf{Finding} & \textbf{Natural-language question} \\
\midrule
\endhead
\midrule
\multicolumn{2}{r}{\small Continued on next page}\\
\endfoot
\bottomrule
\endlastfoot
Medical material & Is medical material present? \\
Arterial wall calcification & Is arterial wall calcification present? \\
Cardiomegaly & Is cardiomegaly present? \\
Pericardial effusion & Is pericardial effusion present? \\
Coronary artery wall calcification & Is coronary artery wall calcification present? \\
Hiatal hernia & Is hiatal hernia present? \\
Lymphadenopathy & Is lymphadenopathy present? \\
Emphysema & Is emphysema present? \\
Atelectasis & Is atelectasis present? \\
Lung nodule & Is lung nodule present? \\
Lung opacity & Is lung opacity present? \\
Pulmonary fibrotic sequela & Is pulmonary fibrotic sequela present? \\
Pleural effusion & Is pleural effusion present? \\
Mosaic attenuation pattern & Is mosaic attenuation pattern present? \\
Peribronchial thickening & Is peribronchial thickening present? \\
Consolidation & Is consolidation present? \\
Bronchiectasis & Is bronchiectasis present? \\
Interlobular septal thickening & Is interlobular septal thickening present? \\
\end{longtable}
\endgroup

\clearpage
\paragraph{Attribute questions.}
These are conditional descriptions and do not establish finding presence.

\begingroup
\small
\setlength{\tabcolsep}{4pt}
\renewcommand{\arraystretch}{1.13}
\begin{longtable}{@{}>{\raggedright\arraybackslash}p{0.24\linewidth}>{\raggedright\arraybackslash}p{0.16\linewidth}>{\raggedright\arraybackslash}p{0.54\linewidth}@{}}
\caption{Attribute queries: all 89 questions.}\label{tab:plan_all_attributes}\\
\toprule
\textbf{Finding} & \textbf{Attribute} & \textbf{Natural-language question} \\
\midrule
\endfirsthead
\multicolumn{3}{l}{\small\itshape Attribute queries: all 89 questions. (continued)}\\
\toprule
\textbf{Finding} & \textbf{Attribute} & \textbf{Natural-language question} \\
\midrule
\endhead
\midrule
\multicolumn{3}{r}{\small Continued on next page}\\
\endfoot
\bottomrule
\endlastfoot
Medical material & appearance & What is the appearance of the medical material? \\
Medical material & location & What is the location of the medical material? \\
Medical material & material type & What is the material type of the medical material? \\
Arterial wall calcification & location & What is the location of the arterial wall calcification? \\
Arterial wall calcification & severity & What is the severity of the arterial wall calcification? \\
Arterial wall calcification & size & What is the size of the arterial wall calcification? \\
Cardiomegaly & location & What is the location of the cardiomegaly? \\
Cardiomegaly & severity & What is the severity of the cardiomegaly? \\
Pericardial effusion & appearance & What is the appearance of the pericardial effusion? \\
Pericardial effusion & location & What is the location of the pericardial effusion? \\
Pericardial effusion & severity & What is the severity of the pericardial effusion? \\
Pericardial effusion & thickness & What is the thickness of the pericardial effusion? \\
Coronary artery wall calcification & affected vessel & What is the affected vessel of the coronary artery wall calcification? \\
Hiatal hernia & location & What is the location of the hiatal hernia? \\
Hiatal hernia & severity & What is the severity of the hiatal hernia? \\
Hiatal hernia & type & What is the type of the hiatal hernia? \\
Lymphadenopathy & location & What is the location of the lymphadenopathy? \\
Lymphadenopathy & morphology & What is the morphology of the lymphadenopathy? \\
Lymphadenopathy & pathological status & What is the pathological status of the lymphadenopathy? \\
Lymphadenopathy & size & What is the size of the lymphadenopathy? \\
Emphysema & associated features & What are the associated features of the emphysema? \\
Emphysema & density & What is the density of the emphysema? \\
Emphysema & location & What is the location of the emphysema? \\
Emphysema & severity & What is the severity of the emphysema? \\
Emphysema & size & What is the size of the emphysema? \\
Emphysema & type & What is the type of the emphysema? \\
Atelectasis & associated signs & What are the associated signs of the atelectasis? \\
Atelectasis & cause & What is the cause of the atelectasis? \\
Atelectasis & distribution & What is the distribution of the atelectasis? \\
Atelectasis & location & What is the location of the atelectasis? \\
Atelectasis & morphology & What is the morphology of the atelectasis? \\
Atelectasis & relation & What is the relation of the atelectasis? \\
Atelectasis & severity & What is the severity of the atelectasis? \\
Atelectasis & shape & What is the shape of the atelectasis? \\
Atelectasis & size & What is the size of the atelectasis? \\
Atelectasis & type & What is the type of the atelectasis? \\
Lung nodule & density & What is the density of the lung nodule? \\
Lung nodule & etiology & What is the etiology of the lung nodule? \\
Lung nodule & location & What is the location of the lung nodule? \\
Lung nodule & margin & What is the margin of the lung nodule? \\
Lung nodule & number & What is the number of the lung nodule? \\
Lung nodule & pattern & What is the pattern of the lung nodule? \\
Lung nodule & shape & What is the shape of the lung nodule? \\
Lung nodule & size & What is the size of the lung nodule? \\
Lung opacity & associated sign & What is the associated sign of the lung opacity? \\
Lung opacity & diagnosis & What is the diagnosis of the lung opacity? \\
Lung opacity & distribution & What is the distribution of the lung opacity? \\
Lung opacity & location & What is the location of the lung opacity? \\
Lung opacity & margin & What is the margin of the lung opacity? \\
Lung opacity & morphology & What is the morphology of the lung opacity? \\
Lung opacity & severity & What is the severity of the lung opacity? \\
Lung opacity & shape & What is the shape of the lung opacity? \\
Lung opacity & size & What is the size of the lung opacity? \\
Pulmonary fibrotic sequela & associated sign & What is the associated sign of the pulmonary fibrotic sequela? \\
Pulmonary fibrotic sequela & etiology & What is the etiology of the pulmonary fibrotic sequela? \\
Pulmonary fibrotic sequela & location & What is the location of the pulmonary fibrotic sequela? \\
Pulmonary fibrotic sequela & morphology & What is the morphology of the pulmonary fibrotic sequela? \\
Pulmonary fibrotic sequela & severity & What is the severity of the pulmonary fibrotic sequela? \\
Pulmonary fibrotic sequela & size & What is the size of the pulmonary fibrotic sequela? \\
Pulmonary fibrotic sequela & subtype & What is the subtype of the pulmonary fibrotic sequela? \\
Pleural effusion & location & What is the location of the pleural effusion? \\
Pleural effusion & morphology & What is the morphology of the pleural effusion? \\
Pleural effusion & quantity & What is the quantity of the pleural effusion? \\
Mosaic attenuation pattern & associated sign & What is the associated sign of the mosaic attenuation pattern? \\
Mosaic attenuation pattern & etiology & What is the etiology of the mosaic attenuation pattern? \\
Mosaic attenuation pattern & location & What is the location of the mosaic attenuation pattern? \\
Mosaic attenuation pattern & severity & What is the severity of the mosaic attenuation pattern? \\
Peribronchial thickening & associated findings & What are the associated findings of the peribronchial thickening? \\
Peribronchial thickening & distribution & What is the distribution of the peribronchial thickening? \\
Peribronchial thickening & location & What is the location of the peribronchial thickening? \\
Peribronchial thickening & severity & What is the severity of the peribronchial thickening? \\
Consolidation & associated sign & What is the associated sign of the consolidation? \\
Consolidation & boundary & What is the boundary of the consolidation? \\
Consolidation & distribution & What is the distribution of the consolidation? \\
Consolidation & location & What is the location of the consolidation? \\
Consolidation & morphology & What is the morphology of the consolidation? \\
Consolidation & severity & What is the severity of the consolidation? \\
Consolidation & shape & What is the shape of the consolidation? \\
Consolidation & size & What is the size of the consolidation? \\
Bronchiectasis & associated sign & What is the associated sign of the bronchiectasis? \\
Bronchiectasis & distribution & What is the distribution of the bronchiectasis? \\
Bronchiectasis & location & What is the location of the bronchiectasis? \\
Bronchiectasis & severity & What is the severity of the bronchiectasis? \\
Bronchiectasis & type & What is the type of the bronchiectasis? \\
Interlobular septal thickening & associated sign & What is the associated sign of the interlobular septal thickening? \\
Interlobular septal thickening & etiology & What is the etiology of the interlobular septal thickening? \\
Interlobular septal thickening & location & What is the location of the interlobular septal thickening? \\
Interlobular septal thickening & morphology & What is the morphology of the interlobular septal thickening? \\
Interlobular septal thickening & severity & What is the severity of the interlobular septal thickening? \\
\end{longtable}
\endgroup

\end{document}